\documentclass[runningheads]{llncs}

\usepackage{eccv}

\usepackage{eccvabbrv}

\usepackage{graphicx}
\usepackage{booktabs}
\usepackage{array}
\usepackage{xcolor}
\usepackage{multirow}
\usepackage{listings}
\usepackage[T1]{fontenc}
\usepackage{subcaption}
\usepackage{fvextra}
\usepackage{makecell}
\usepackage{seqsplit}
\usepackage{listings}
\usepackage{fvextra}
\usepackage[accsupp]{axessibility}  

\usepackage{hyperref}

\usepackage{orcidlink}

\begin{document}

\title{SpecVQA: A Benchmark for Spectral Understanding and Visual Question Answering in Scientific Images} 


\titlerunning{SpecVQA}


\author{Jialu Shen \and
Han Lyu \and
Suyang Zhong \and
Hanzheng Li \and
Haoyi Tao \and
Nan Wang \and
Changhong Chen \and
Xi Fang{\thanks{Corresponding author}}}

\authorrunning{Shen et al.}

\institute{DP Technology }

\maketitle

\begin{abstract}
Spectra are a prevalent yet highly information-dense form of scientific imagery, presenting substantial challenges to multimodal large language models (MLLMs) due to their unstructured and domain-specific characteristics.
Here we introduce \textbf{SpecVQA}, a professional scientific-image benchmark for evaluating multimodal models on scientific spectral understanding, covering 7 representative spectrum types with expert-annotated question–answer pairs. 
The aim comprises two aspects: spectra scientific QA evaluation and corresponding underlying task evaluation. 
SpecVQA contains 620 figures and 3100 QA pairs curated from peer-reviewed literature, targeting both direct information extraction and domain-specific reasoning. 
To effectively reduce token length while preserving essential curve characteristics, we propose a spectral data sampling and interpolation reconstruction approach. 
Ablation studies further confirm that the approach achieves substantial performance improvements on the proposed benchmark.
We test the capability of prominent MLLMs in scientific spectral understanding on our benchmark and present a leaderboard. 
This work represents an essential step toward enhancing spectral understanding in multimodal large models and suggests promising directions for extending visual–language models to broader scientific research and data analysis.
\end{abstract}

\section{Introduction}
\label{sec:intro}

\begin{figure*}[h]
    \centering
    \makebox[\textwidth]{\includegraphics[width=1.05\textwidth]{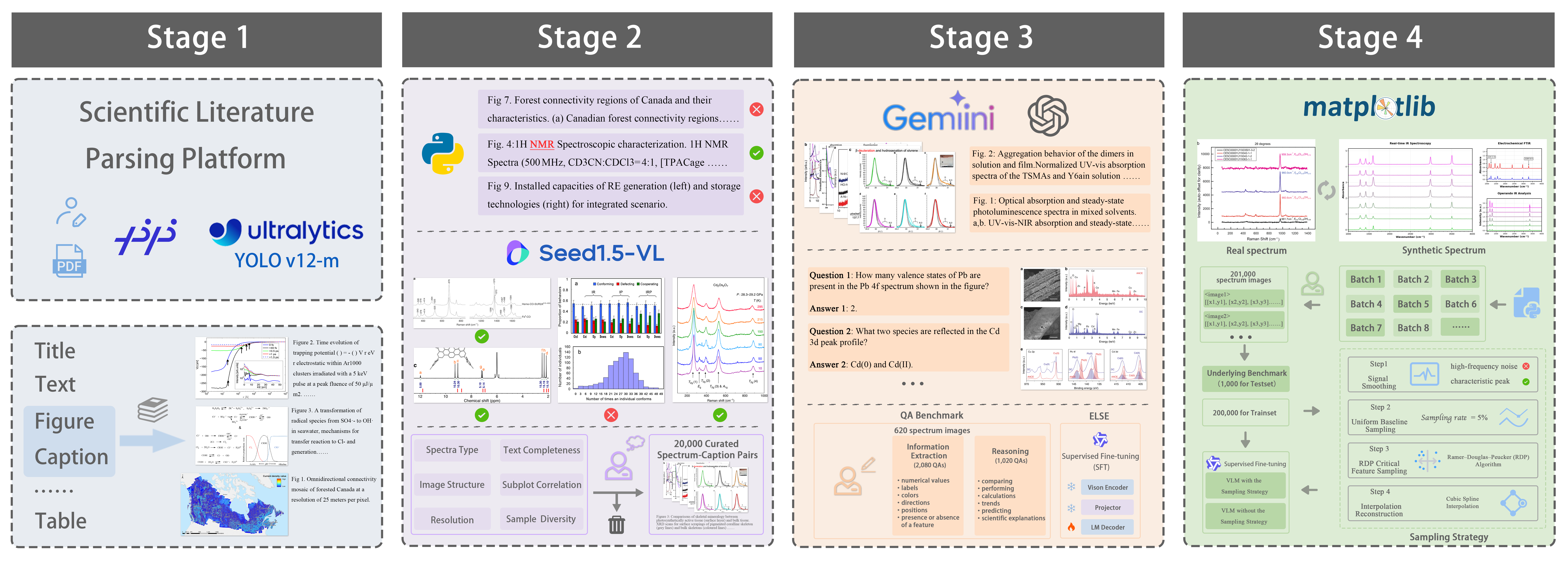}}
    \caption{Overview of data curation and the proposed SpecVQA benchmark. Stage 1 automatically harvests high-quality figure–caption pairs from peer-reviewed journals. Stage 2 performs multi-stage filtering and spectrum-type classification, yielding a curated set of 20k spectrum–caption pairs under expert supervision. Stage 3 leverages MLLMs to distill QA pairs from the curated figures; domain experts then re-annotate representative cases and assign them to two semantic categories. Stage 4 synthesizes an additional 201k  underlying samples to support model evaluation and to assess the effectiveness of our sampling strategy.}
    \label{fig:placeholder}
\end{figure*}

The emergence of Multimodal Large Language Models (MLLMs) has marked a new milestone in artificial intelligence, demonstrating remarkable advancements in visual–language understanding, cross-modal reasoning, and zero-shot generalization across diverse domains \cite{radford2021learning, changpinyo2021conceptual, liu2023visual}. Both proprietary systems such as GPT-5 \cite{openai_gpt5_2025}, Gemini-3 \cite{gemini3pro}, Claude-4.5 \cite{claude45sonnet} and advanced open-source architectures such as Qwen3-VL \cite{bai2025qwen3vltechnicalreport} exemplify an increasingly unified understanding of the world, continuously pushing the boundaries of machine perception \cite{yin2024survey, roberts2024scifibench}. The potential to translate this broad, general intelligence into domain-specific scientific discovery is substantial, particularly in fields that depend heavily on complex data visualization. Despite strong general-purpose reasoning abilities, current MLLMs still struggle with scientific spectral figures, where dense curves, precise numerical reading, and domain knowledge are tightly coupled. It requires a synthesis of pattern recognition and deep mechanistic understanding \cite{houhou2021trends, fine2020spectral}.

In spectral image understanding, MLLMs typically target two complementary tasks:  Scientific QA Task and Underlying Construction Task. \\
\textbf{(1) Scientific QA Task:} expert-level interpretation and reasoning based solely on the visual characteristics of the spectral plot and the domain knowledge implicitly encoded within it. The fundamental limitations of current MLLMs in spectral understanding are summarized as follows.
\begin{itemize}
    \item Mislocalization of Fine Features. Inability to accurately locate or distinguish subtle spectral details, such as minor impurity peaks, small chemical shifts, or the precise positions of crystalline reflections.
    \item Inaccurate Quantitative Extraction. Errors in reading key numerical values from axes or curves, including integration ratios, maximum absorption wavelengths ($\lambda_{\text{max}}$), and precise binding energies.
    \item Flawed Mechanistic Reasoning. Logical inconsistencies when connecting visual evidence to scientific principles, resulting in incorrect structure elucidation, phase identification, or interpretation of fragmentation pathways.
\end{itemize}
\textbf{(2) Underlying Construction Task:} extracting and reconstructing the high-density numerical data points that define the spectral curve, often for a specific sub-figure within a composite visualization. 
Relative early works include ChartOCR \cite{luo2021chartocr} and Plot2Spectra \cite{jiang2022plot2spectra}. 
This task is closely related to chart data extraction and plot digitization, but differs in that it focuses on dense spectral curve reconstruction from sub-figures embedded in composite scientific visualizations, requiring accurate curve localization, separation, and interpolation. 

We show some bad cases of these two tasks in the Appendix.

In the scientific QA task, it is noted that these limitations cannot be accurately assessed or diagnosed by existing benchmarks. Standard VQA datasets \cite{goyal2017making} primarily evaluate general object recognition, whereas chart-based VQA datasets \cite{methani2020plotqa, masry2022chartqa} focus on elementary data retrieval without requiring scientific knowledge grounding. 
To address this gap, we propose SpecVQA (Spectral Visual Question Answering)\footnote{https://huggingface.co/datasets/UniParser/SpecVQA}, a professional scientific-image benchmark for evaluating multimodal models on scientific spectral understanding. It encompasses the seven most prevalent and analytically important types of spectra, which focus enables a quantitative evaluation of MLLMs in terms of both scientific accuracy and their capacity to leverage embedded domain knowledge, and thereby guiding the development of more robust, domain-aware multimodal models.

In the underlying task, high-resolution spectral data present a major token-efficiency challenge for MLLMs, especially in tasks requiring reconstruction or regression of the underlying curves. Unlike general charts with sparse key points, spectra comprise thousands of data points, each representing critical fluctuations that models are required to capture with high accuracy. 
Whether learning from or predicting these points, MLLMs will tokenize them into a longer sequence compared with those in other tasks, which might impose heavy computational and memory burdens. Attempts to compress the data often lead to truncation or oversimplification and making fine-tuning on large spectral datasets impractical. This challenge arises regardless of input image length, as it depends on the minimum token count required to faithfully represent the spectral curve.

To overcome the underlying task challenges, we introduce an efficient data sampling and interpolation reconstruction strategy. This method aims to drastically reduce output token length while preserving the spectral features, alleviating the token bottleneck and enabling large-scale MLLMs fine-tuning on high-density scientific data. The resulting test set also forms part of our benchmark, which it used to quantitatively evaluate the performance of several large models on the spectral underlying task.
In summary, this paper makes the following major contributions:\\
\textbf{A High-quality Benchmark (SpecVQA).} We introduce a comprehensive, domain-specific VQA benchmark rigorously curated by scientific experts, specifically targeting the seven most critical types of scientific spectra.\\
\textbf{An Efficient Data Sampling Strategy.} We propose Efficient Data Sampling and Interpolation Reconstruction strategy that employs adaptive sampling and interpolation to substantially reduce text token length, improving training efficiency while preserving spectral fidelity.\\
\textbf{A Comprehensive Leaderboard.} We establish a detailed leaderboard that quantitatively evaluates and compares major MLLMs on the SpecVQA benchmark, identifying their limitations and providing actionable insights for future domain-specialized multimodal model development.

\section{Related Work}
\label{sec:review}

Our work is positioned at the intersection of general MLLMs, specialized visual question answering, and advanced computational methods for scientific data.

\subsection{Foundations of MLLMs}

The foundation of MLLMs is the integration of powerful vision encoders with large language models (LLMs) \cite{vaswani2017attention, dosovitskiy2020image}. Early approaches demonstrated the feasibility of aligning visual and linguistic feature spaces through contrastive learning \cite{radford2021learning, li2021align}, while subsequent models introduced sophisticated projection layers to bridge the gap between frozen image encoders and LLMs \cite{li2023blip, alayrac2022flamingo}.
The recent surge of instruction-tuned MLLMs, such as LLaVA \cite{liu2023visual, li2024llava} and Qwen-VL \cite{bai2023qwenvl, bai2025qwen2}, has further advanced zero-shot and few-shot reasoning capabilities across a wide range of multimodal tasks \cite{zhou2020unified, tan2019lxmert}. Trained on massive web-scale corpora \cite{schuhmann2022laion}, these models exhibit remarkable generalization ability. However, such generalized training inherently dilutes model sensitivity to the highly specialized, formalized, and quantitative visual semantics found in scientific spectra \cite{roberts2024scifibench}.

\subsection{Specialized VQA and Scientific Reasoning}

As general VQA benchmarks mature, research attention has increasingly shifted toward specialized visual reasoning.\\
\textbf{Chart and Data Visualization VQA.} Efforts in data visualization understanding such as PlotQA \cite{methani2020plotqa} and ChartQA \cite{masry2022chartqa} focus on interpreting standard business and statistical charts (e.g., bar, line, and pie charts). While these tasks involve axis reading and arithmetic comparison, they primarily deal with discrete or low-density data points and emphasize simple data retrieval rather than scientific interpretation. They fall short of supporting advanced reasoning, such as distinguishing $\pi$–$\pi^*$ from $\text{n}$–$\pi^*$ electronic transitions in UV–Vis spectra or inferring chemical environments from peak multiplicity in NMR spectra. The specialized nature of spectral interpretation thus demands a distinct category of reasoning beyond elementary data extraction.\\
\textbf{Scientific Document and Diagram Understanding.} Broader scientific benchmarks such as ScienceQA \cite{saikh2022scienceqa} and other datasets for figure understanding, integrate textual and visual modalities. However, their visual elements are typically schematic or illustrative (e.g., flowcharts or simple diagrams), lacking the continuous, high-density data and strict analytical constraints that characterize experimental spectra. Our work is distinguished by its requirement for expert-level quantitative analysis and knowledge-driven deduction applied directly to images of raw scientific measurements.

\subsection{Computational Methods for Spectral Data and Long Context VLM}

\textbf{Traditional Spectral Analysis.} Spectral analysis has long been a cornerstone of computational spectroscopy and chemometrics. Techniques ranging from Partial Least Squares (PLS) to modern deep learning models (CNNs and RNNs) have been employed on raw numerical data for tasks such as classification, compound identification, and property prediction \cite{houhou2021trends, fine2020spectral, lansford2020infrared}. These approaches serve as predictive tools operating purely on numerical inputs, which fundamentally differ from our goal of achieving visual language grounding based on spectral images. They do not address the need to bridge MLLMs’ visual perception with scientific language understanding.\\
\textbf{Data Sampling and Dimensionality Reduction.} Efficiently processing the high-density data generated by modern instruments poses a longstanding challenge. Reducing the number of sampled points while retaining essential curve characteristics is critical for computational efficiency. Classical simplification methods such as the Ramer–Douglas–Peucker (RDP) algorithm \cite{ramer1972iterative, douglas1973algorithms} achieve this through adaptive piecewise linear approximation, whereas more recent approaches leverage Topological Data Analysis (TDA) \cite{wasserman2018topological} and persistent homology to capture scale-invariant geometric features. Although these techniques are well-established in signal processing and computer graphics, their targeted integration into multimodal large-model pipelines to alleviate the long-tokenization burden of high-fidelity scientific curves remains unexplored.

\section{Method}
\label{sec:method}

To address the token length crisis resulting from high-density spectral data, we propose the Data Sampling and Interpolation Reconstruction Strategy. This technique significantly reduces the length of the output text tokens without compromising the scientific fidelity required for accurate interpretation.

\begin{figure}[t]
\centering
\begin{minipage}{0.48\textwidth}
    \centering
    \includegraphics[width=\textwidth]{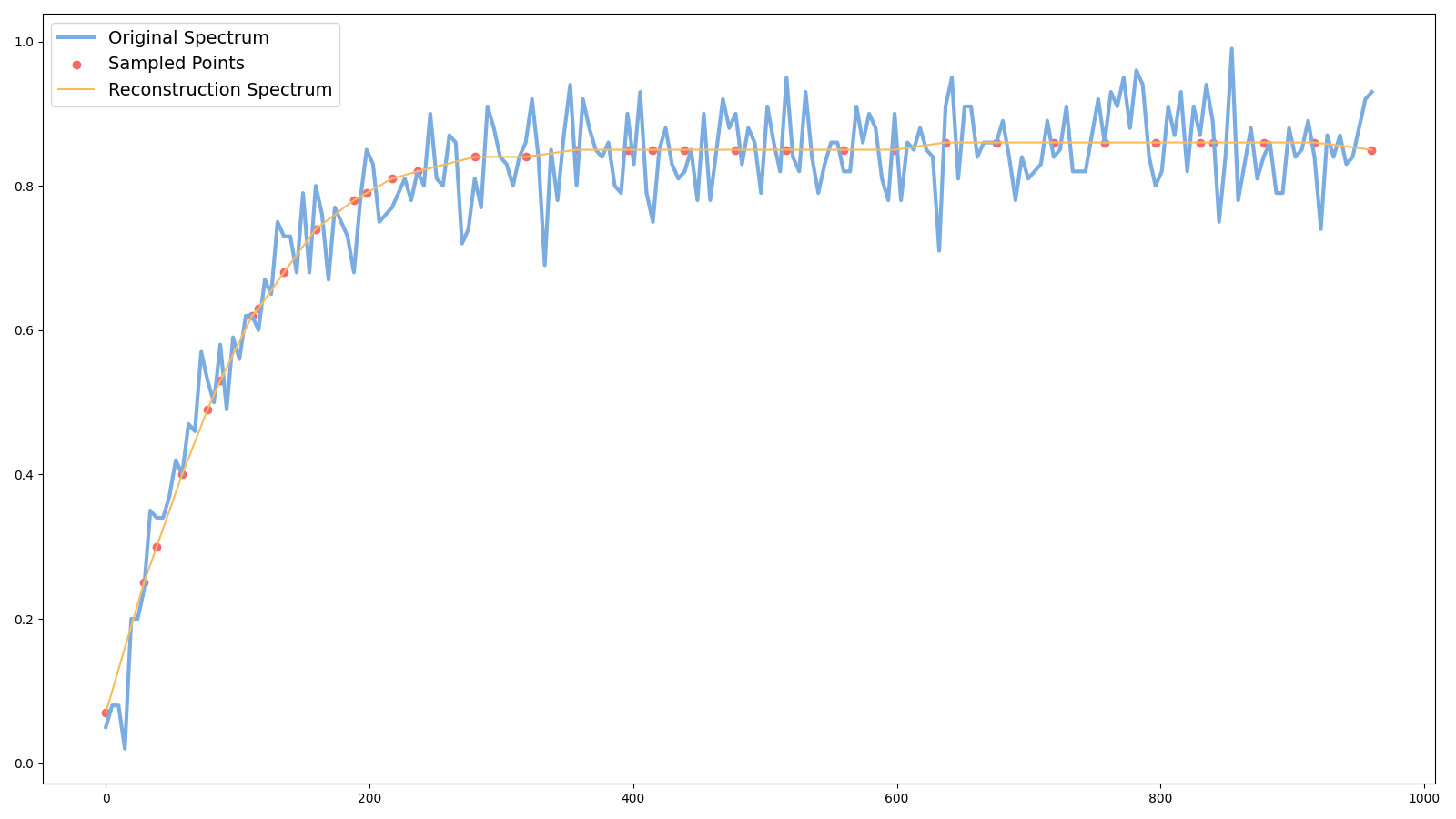}
    \subcaption{}
\end{minipage}
\hfill
\begin{minipage}{0.48\textwidth}
    \centering
    \includegraphics[width=\textwidth]{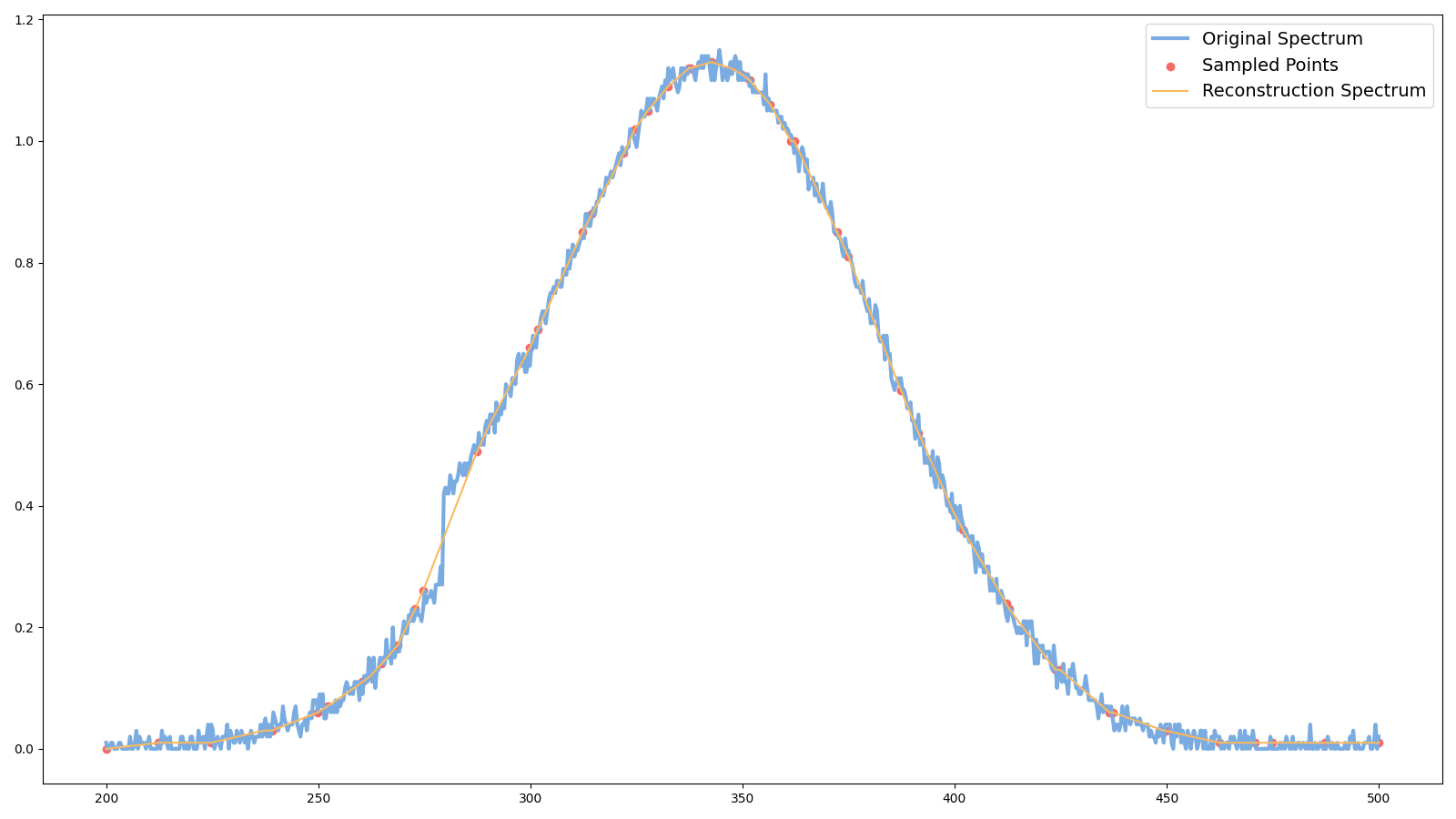}
    \subcaption{}
\end{minipage}
\hfill
\begin{minipage}{0.48\textwidth}
    \centering
    \includegraphics[width=\textwidth]{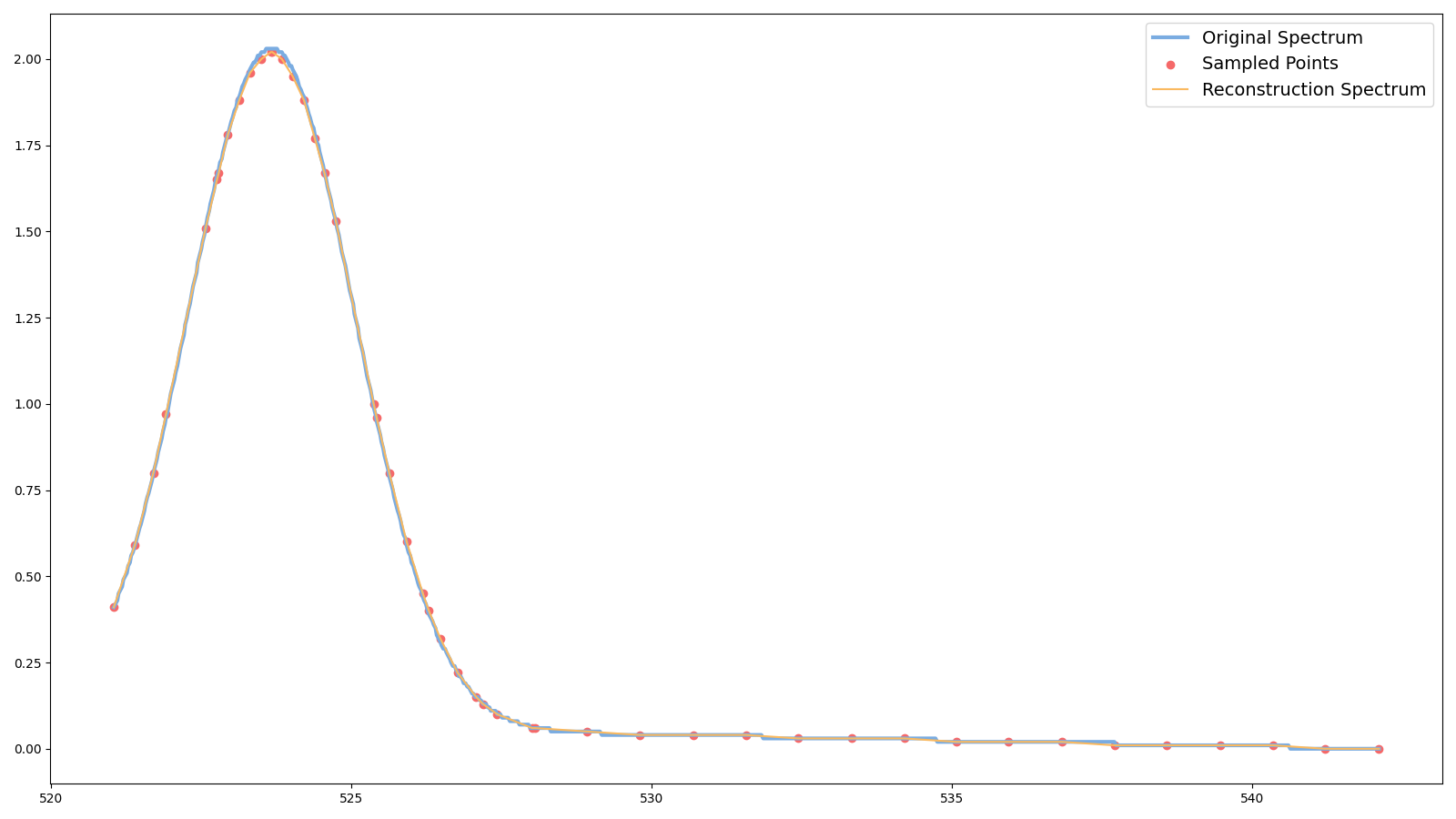}
    \subcaption{}
\end{minipage}
\hfill
\begin{minipage}{0.48\textwidth}
    \centering
    \includegraphics[width=\textwidth]{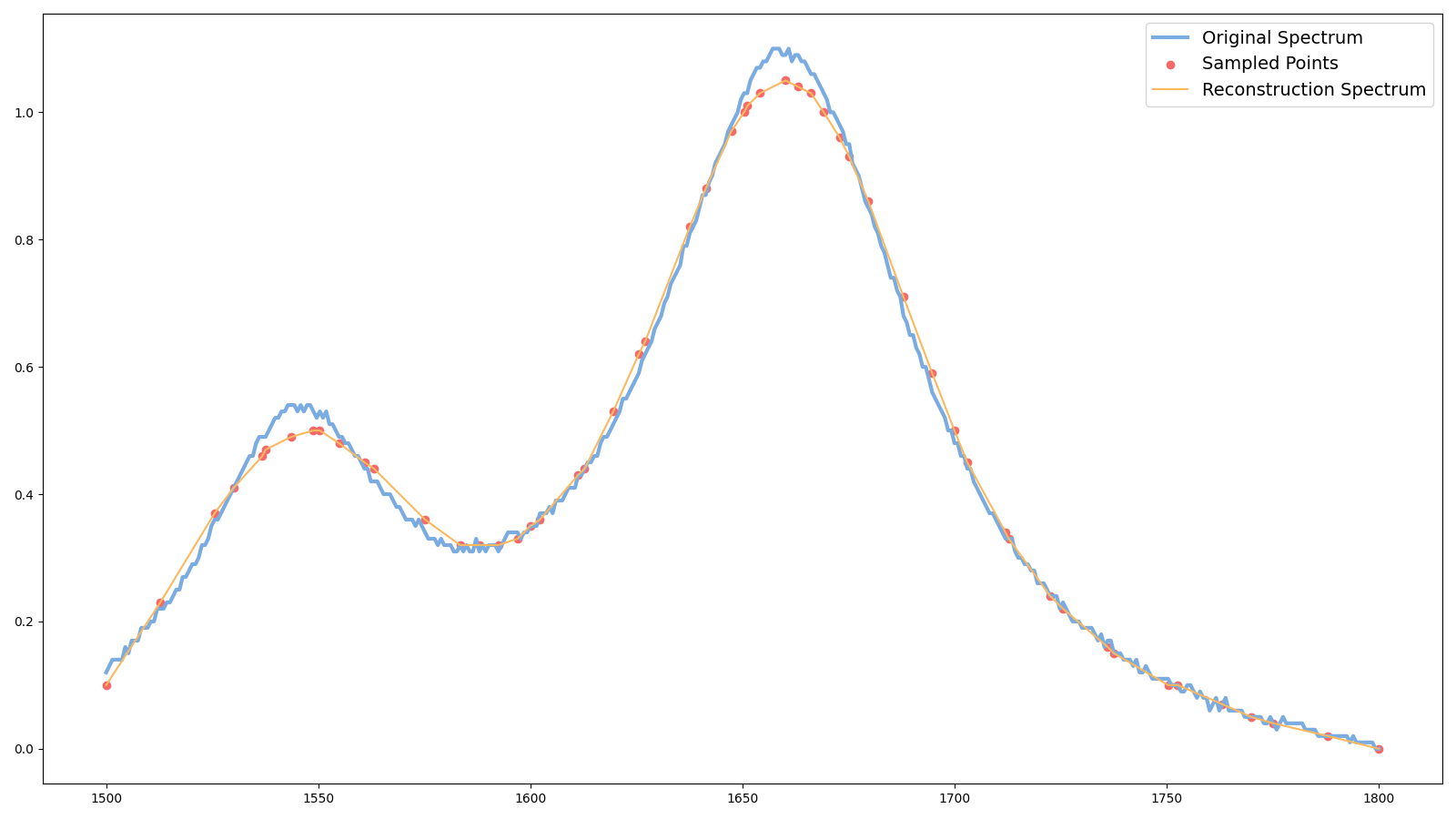}
    \subcaption{}
\end{minipage}
\caption{Comparison of the noisy real data curve, sampled points, and the reconstructed smooth curve. Four examples from the experiments are presented.}
\label{fig:rec}
\end{figure}

\subsection{Adaptive Key-point Sampling}

Unlike uniform downsampling which risks discarding narrow and critical signal peaks, our method employs a three stage strategy to maximize the preservation of scientific features while achieving a high reduction ratio $R$. The original high-resolution spectral curve is defined by $N$ data points: $\mathcal{S}_{\text{ori}} = \{(x_i, y_i)\}_{i=1}^N$.\\
\textbf{Step 1: Signal Smoothing.} The raw data is first subjected to a Savitzky–Golay (S-G) filter \cite{savitzky1964smoothing} to effectively reduce high-frequency noise without distorting the characteristic peak shapes and widths. This preprocessing step enhances the robustness of subsequent feature extraction.\\
\textbf{Step 2: Uniform Baseline Sampling.} To ensure coverage of baseline and low-feature regions, we uniformly sample approximately $5\%$ of the points from the smoothed curve. This guarantees a minimal token representation.\\
\textbf{Step 3: Critical Feature Sampling.} We apply the RDP algorithm \cite{ramer1972iterative, douglas1973algorithms} to the smoothed curve. RDP is a line simplification algorithm that recursively identifies and keeps only the most crucial points required to approximate the curve within a specified maximum distance threshold $\epsilon$. After this step, the sampled points inherently encode the most significant bends and features of the spectral curve, preserving peak summits and critical changes in slope. The final sampled set $\mathcal{S}_{\text{sampled}}$ is the union of points from step 2 and step 3.\\

\subsection{Interpolation Reconstruction}
The sparse sampled point set $\mathcal{S}_{\text{sampled}}$ is then used to reconstruct a new smooth visual representation $I_{\text{recovered}}$. We employ cubic spline interpolation since it ensures second-order continuity, which is crucial for preserving the natural shape and smoothness of spectral curves that MLLMs must interpret.
\begin{equation}
I_{\text{recovered}} = \text{Image}(\text{CubicSpline}(\mathcal{S}_{\text{sampled}}))
\end{equation}
The resulting $I_{\text{recovered}}$ reconstruct the original data line well and drastically shortening the input token length, thus mitigating memory and truncation issues. Figure \ref{fig:rec} shows the comparsion of real data line and reconstruction data line.

\subsection{Fidelity Evaluation}
To quantitatively evaluate the efficiency of our strategy in preserving the spectral curve's shape and fidelity after downsampling, we employ several metrics.\\ 
\textbf{Chamfer Distance ($d_{CD}$)} \cite{fan2017point, chazelle1985optimal} measures the average squared Euclidean distance between the nearest points of two point sets. Let $C_{\text{true}}$ denote the original high-resolution point cloud and $C_{\text{rec}}$ denote the reconstructed point cloud obtained from sampled points. A lower value of $d_{CD}$ indicates better preservation of geometric fidelity between the reconstructed and original point clouds.
\begin{equation*}
d_{CD}\left(C_{\text {true }}, C_{\text {rec }}\right) = \frac{1}{\left|C_{\text {true }}\right|} \sum_{p \in C_{\text {true }}} \min _{q \in C_{\text {rec }}}\|p-q\|^2
+\frac{1}{\left|C_{\text {rec }}\right|} \sum_{q \in C_{\text {rec }}} \min _{p \in C_{\text {true }}}\|q-p\|^2.
\end{equation*}
\textbf{Hausdorff Distance ($d_{HD}$)} \cite{huttenlocher2002comparing, dubuisson1994modified} measures the maximum distance from a point in one point set to its nearest neighbor in the other set. 
We define $C_{\text{true}}$ as the original high-resolution point cloud and $C_{\text{rec}}$ as the reconstructed point cloud from sampled points. 
A lower $d_{HD}$ indicates that the worst-case geometric deviation between the reconstructed and original point clouds is smaller, implying better structural preservation.
\begin{equation*}
d_{HD}\left(C_{\text {true }}, C_{\text {rec }}\right)= \max \{\max _{p \in C_{\text {true }}} \min _{q \in C_{\text {rec }}}\|p-q\|, \max _{q \in C_{\text {rec }}} \min _{p \in C_{\text {true }}}\|q-p\|\}.
\end{equation*}\\
\textbf{Wasserstein Distance ($d_{WD}$)} 
\cite{villani2009optimal, rubner2000earth} measures the minimal transportation cost required to transform one point distribution into another. 
We define $C_{\text{true}}$ as the original high-resolution point cloud and $C_{\text{rec}}$ as the reconstructed point cloud from sampled points. 
A lower $d_{WD}$ indicates that the reconstructed point cloud better matches the overall spatial distribution of the original point cloud.
\begin{equation*}
d_{WD}\left(C_{\text {true}}, C_{\text {rec}}\right)= \frac{1}{N}\sum_{i=1}^N\|p_i-q_i\|.
\end{equation*}

\section{The SpecVQA Benchmark}
\label{sec:benchmark}
To address the limitations of general MLLMs in interpreting scientific figures and to provide a systematic evaluation framework, we introduce SpecVQA (Spectral Visual Question Answering). SpecVQA is designed to emulate the complex, visually grounded, and knowledge-integrated reasoning process that human experts employ when interpreting experimental spectral data.

\subsection{Benchmark Scope}
\label{sec:scope}
SpecVQA focuses on seven widely utilized types of spectra frequently encountered in scientific research. This specialization ensures that the benchmark’s evaluation results are directly relevant to practical scientific analysis. The spectrum types and their core interpretation requirements are as Table \ref{tab:type}.

\begin{table}[h]
\centering
\caption{The seven spectrum types and their core interpretation requirements}
\label{tab:type}
\begin{tabular}{|>{\centering\arraybackslash}m{4cm}|>{\centering\arraybackslash}m{7.5cm}|}
\hline
\textbf{Spectrum Type} & \textbf{Description} \\
\hline
NMR (Nuclear Magnetic Resonance) & Involves interpreting chemical shifts ($\delta$), integration areas, coupling constants ($J$), and multiplicity to deduce molecular structures. \\
\hline
IR (Infrared Absorption Spectroscopy) & Involves identifying characteristic vibrational frequencies ($\text{cm}^{-1}$) to determine the presence or absence of specific functional groups. \\
\hline
XRD (X-ray Diffraction) & Requires analyzing diffraction peak positions ($2\theta$), intensities, and widths to infer crystal structures, lattice parameters, and crystallite sizes. \\
\hline
Raman (Raman Spectroscopy) & Focuses on interpreting Raman-active vibrational modes to probe molecular symmetry and bonding characteristics, particularly in materials such as carbon-based systems. \\
\hline
MS (Mass Spectrometry). & Involves analyzing molecular ion peaks and fragmentation patterns to determine molecular weight and identify structural fragments. \\
\hline
UV-Vis (Ultraviolet-Visible Spectrophotometry). & Requires extracting maximum absorption wavelengths ($\lambda_{\text{max}}$) and quantifying concentrations based on absorption intensity. \\
\hline
XPS (X-ray Photoelectron Spectroscopy). & Involves analyzing core-level peak binding energies to determine elemental composition and chemical states at the sample surface. \\
\hline
\end{tabular}
\end{table}

\subsection{Data Curation}

\subsubsection{High-Quality Data Acquisition.}
\label{sec:data_acquisition}

SpecVQA is constructed from high-quality, real-world scientific data, emphasizing both accessibility and authoritative provenance. The raw image corpus is primarily aggregated from Open Access (OA) literature, the arXiv preprint server, and large-scale technical reports. Refer to \textbf{Uniparser} \cite{fang2025uni} and \textbf{OmniScience} \footnote{https://huggingface.co/datasets/UniParser/OmniScience} for the process of acquiring literature data.
Specifically, to ensure the data authority and impact, we prefer figures extracted from papers with high citation counts or published in high-impact factor journals. We will give more consideration to literature from the last 20 years to ensure image clarity, while also ensuring the integrity of the dataset by not omitting classic literature in related fields.

\subsubsection{Systematic Image Segmentation Pipeline.}

The figure corpus was systematically extracted using a dedicated pipeline.  Initial figure-text pairs were extracted coarsely from the literature corpus using keywords corresponding to the seven spectrum types. Then the article provenance including journal and discipline was cross-referenced to perform a layered case extraction, ensuring a diverse, broad coverage of domain knowledge within the chemical and material sciences. 
Image segmentation and extraction are performed the same as OmniScience.

\subsubsection{Refined Figure-Label Pair Selection Rules.}

During the final manual and semi-automated refinement, the following rules were applied to ensure semantic independence and data quality. Of the 60k figure-caption obtained, 30k were automatically selected by regularization method and Seed1.5-VL, and then 20k spectrum-label pairs were retained through manual screening.

\begin{itemize}
    \item Sufficient Description. Only figures where the accompanying text description was sufficient to convey the essential physical or chemical information were retained. Text requiring excessive external context or providing minimal description was discarded.
    \item Single-Type Spectrum. If the figure contained only one spectrum and the text description was complete and context-independent, the figure-text pair was saved, typically named "filename + A".
    \item Comparison Cases. If the figure consisted of multiple sub-figures of the same type, and the text explicitly emphasized their contrast or comparison, the entire original figure was retained without splitting, supporting comparative analysis tasks.
    \item Resolution Consideration. Recognizing the prevalence of low-resolution images in real-world applications, we did not limit selection to only high-resolution spectra, thus ensuring the high practical generalization of the dataset.
    \item Semantic Independence. A core consideration during selection was ensuring that different spectrum sub-figures exhibit weak association, thus ensuring the semantic independence of individual samples.
\end{itemize}

\subsection{QA Benchmark}

\subsubsection{Expert screening.}

To ensure the scientific validity and representativeness of the benchmark, a team of domain experts manually curated a subset of 620 spectral figures from an initial pool of over 20k candidates collected from peer-reviewed journals and open-access scientific databases. After taking into account both time cost and answer quality, all candidate images were used to generate QA pairs via distillation using GPT-o4-mini and Gemini-2.5-Pro, with five QA pairs produced for each image. 

The selection process was guided by following criteria.
Figures were chosen to cover a broad range of visual configurations, including single-spectrum plots, multi-panel figures, and both homogeneous and heterogeneous subplot arrangements, enabling evaluation of MLLMs on complex figure layouts and visual hierarchies. Each figure was also screened to ensure that the accompanying captions or contextual text contained sufficient experimental and interpretive information to support grounded question–answer generation. In addition, experts examined the semantic relationships among subplots, distinguishing between independent and tightly coupled panels to facilitate multi-instance reasoning. The dataset further emphasizes diversity across journals, research fields, and data sources to reduce bias and improve generalizability. Finally, both high and low resolution images were included to reflect practical scenarios and evaluate the visual robustness of MLLMs.

During the selection process, the QA pairs distilled from the chosen images were also reviewed by experts. In addition to verifying factual correctness, the experts were required to assess the semantic relevance of each QA pair. We expected the questions and answers to reflect issues that are genuinely of interest in their research and pose challenges to MLLMs, rather than trivial or nonsensical visual Q\&As. If any distilled pair failed to meet these criteria, the experts revised it until it satisfied the requirements.

Through this expert screening pipeline, SpecVQA achieves a carefully balanced collection of spectral figures that combines scientific rigor, visual diversity, and contextual completeness—forming a reliable foundation for multimodal evaluation in specialized scientific reasoning.

\subsubsection{Multi-Stage Bias-Averse QA Generation.}
\label{sec:qa_generation}

The construction of the SpecVQA question-answering pairs employs a robust, multi-stage pipeline designed to maximize question diversity, minimize MLLM generation bias, and ensure scientific accuracy. The final 3,100 QA pairs are classified into two critical categories based on the required cognitive effort. We provided both Chinese and English version to test the scientific performance of the large model in different languages.
\begin{itemize}
    \item \textbf{Category 1 (L0): Descriptive
Question (2,080 QAs).} This category evaluates the ability to directly understand and interpret visual information in spectral figures. It includes tasks such as extracting textual elements (e.g., titles, labels, legends, and axis information), identifying key values like maxima or minima and peak locations, recognizing patterns or entities that satisfy specific conditions, understanding the layout of multi-panel subplots, and classifying visual features such as peak shapes.

\item \textbf{Category 2 (L1): Reasoning Question (1,020 QAs).} This category focuses on higher-level analytical and reasoning abilities based on the visual content of the figures. It involves comparing multiple entities to draw conclusions, counting elements that meet certain criteria, performing calculations on numerical information in the figure, analyzing trends or predicting changes in spectral patterns, and conducting causal analysis to interpret the scientific phenomena reflected in the data.
\end{itemize}
\textbf{Dataset Statistics.} This part of SpecVQA benchmark comprises 620 unique scientific spectral figures. Through rigorous expert-level generation and filtering, we constructed a total of 3,100 high-quality Question-Answer pairs (an average of 5 QA pairs per figure). The underlying, raw numerical data associated with these spectral images is made publicly available as a part of the benchmark.
The critical distinction of SpecVQA is that it requires models to interpret the spectrum rather than merely read the graph, specifically addressing the high data density and expert knowledge required for curve-based scientific analysis.

\begin{figure}
    \centering
    \includegraphics[width=0.8\linewidth]{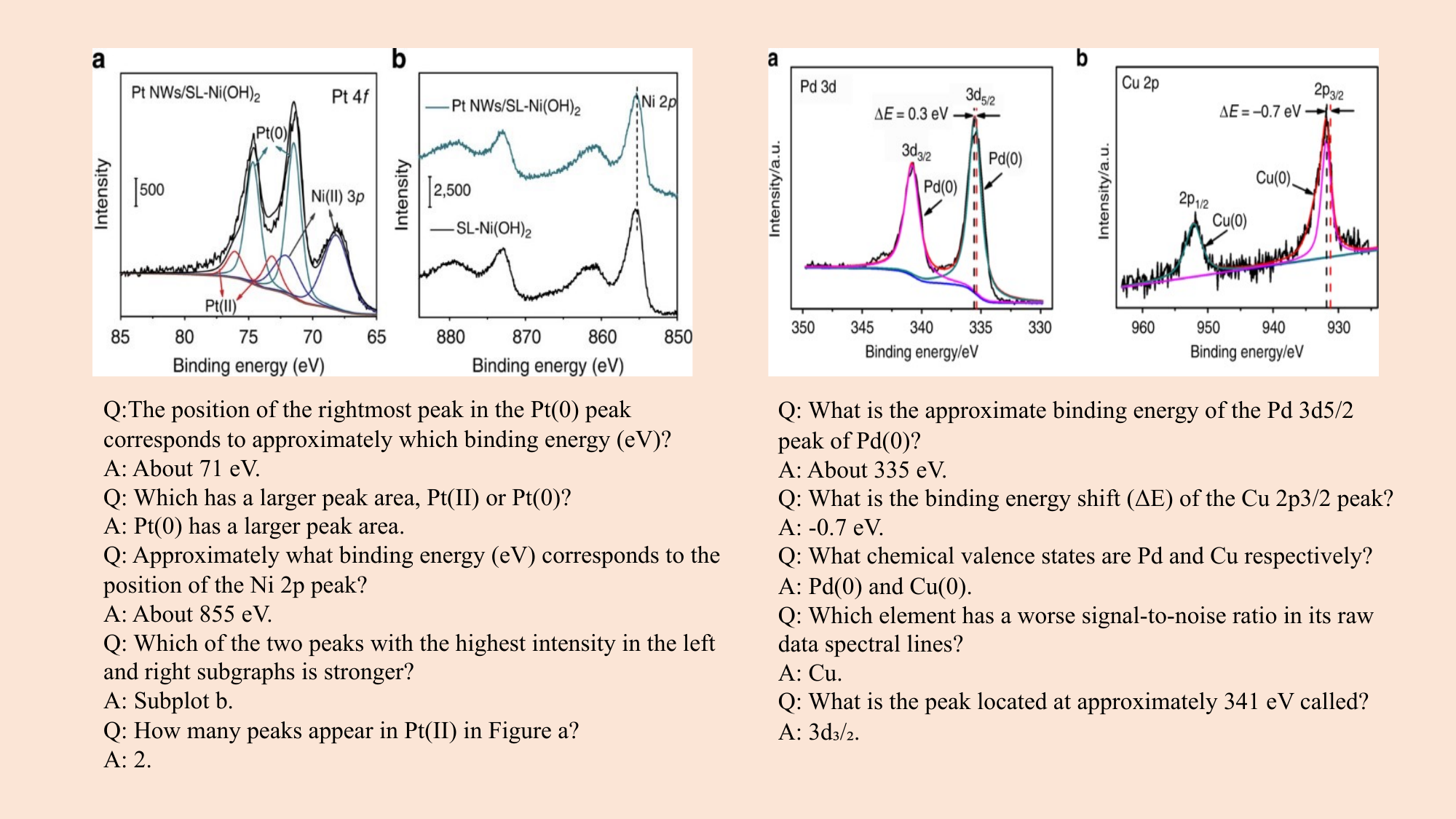}
    \caption{Two examples from the Scientific QA part of the SpecVQA benchmark. Each example contains one spectral figure and five QA pairs in different languages. Here we show the English QA pairs.}
    \label{fig:placeholder}
\end{figure}

\subsection{Underlying Benchmark Construction}
\label{sec:sf-dr_method}

\begin{figure}[t]
\centering
\begin{minipage}{0.35\linewidth}
    \centering
\includegraphics[height=2.5cm,keepaspectratio]{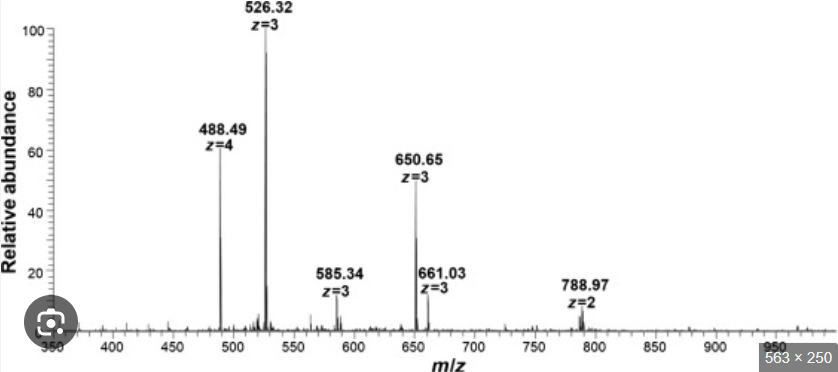}
    \subcaption{ms-real}
\end{minipage}
\hfill
\begin{minipage}{0.31\linewidth}
    \centering
\includegraphics[height=2.5cm,keepaspectratio]{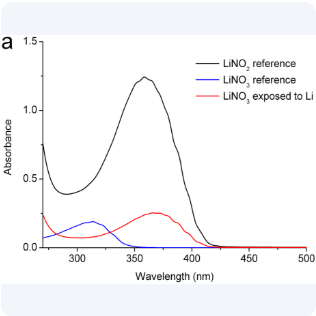}
    \subcaption{uv-real}
\end{minipage}
\hfill
\begin{minipage}{0.31\linewidth}
    \centering
\includegraphics[height=2.5cm,keepaspectratio]{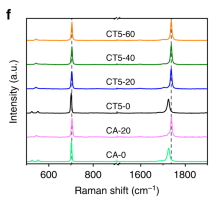}
    \subcaption{raman-real}
\end{minipage}

\begin{minipage}{0.35\linewidth}
    \centering
\includegraphics[height=2.5cm,keepaspectratio]{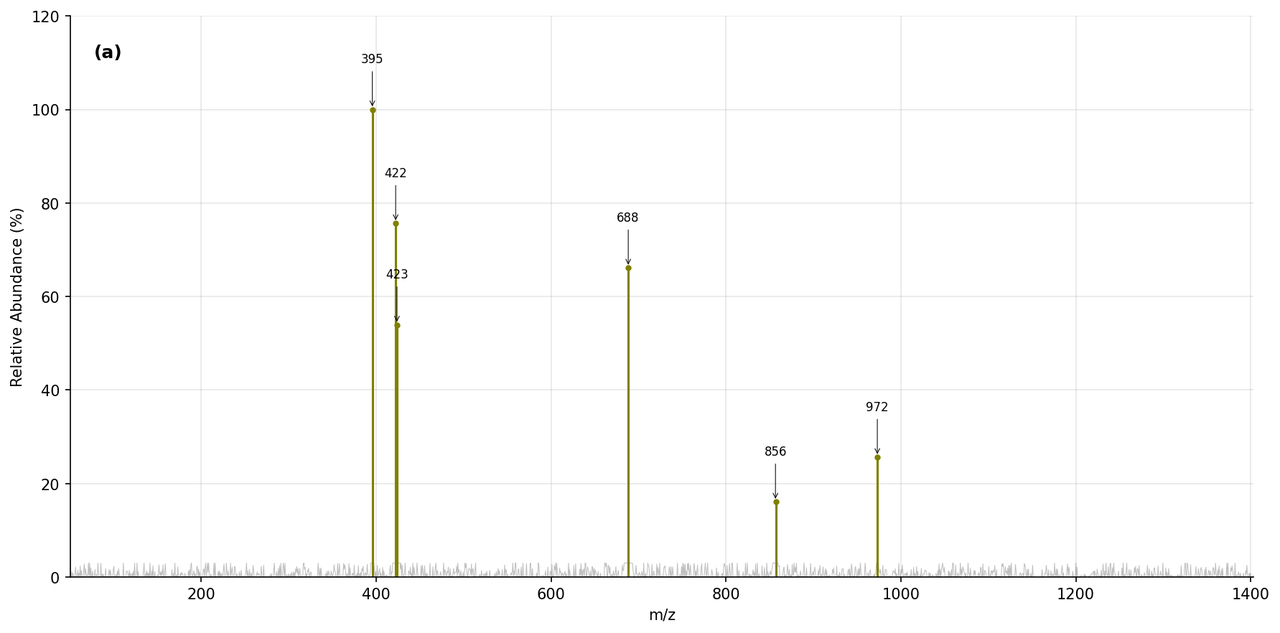}
    \subcaption{ms-generated}
\end{minipage}
\hfill
\begin{minipage}{0.31\linewidth}
    \centering
\includegraphics[height=2.5cm,keepaspectratio]{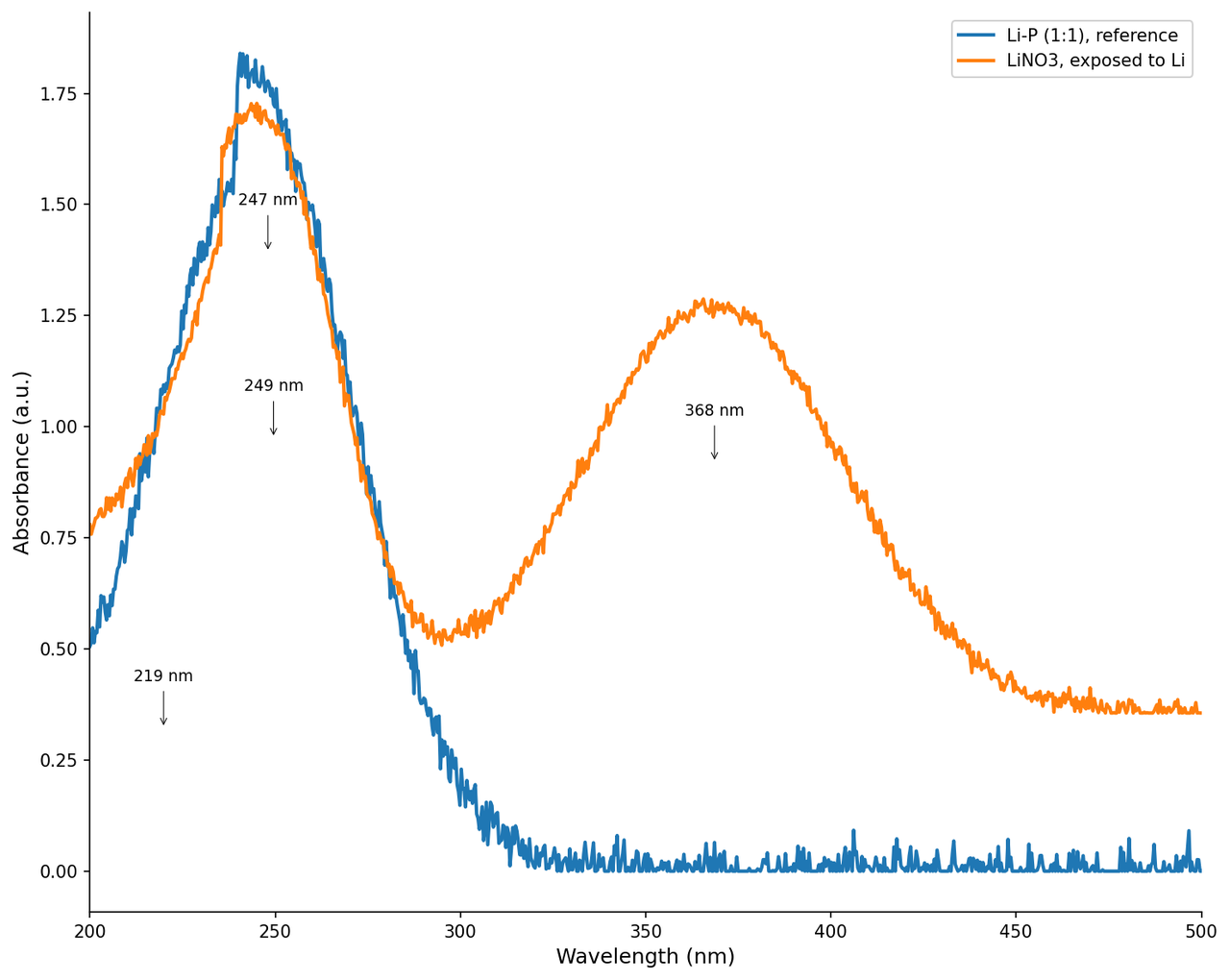}
    \subcaption{uv-generated}
\end{minipage}
\hfill
\begin{minipage}{0.31\linewidth}
    \centering
\includegraphics[height=2.5cm,keepaspectratio]{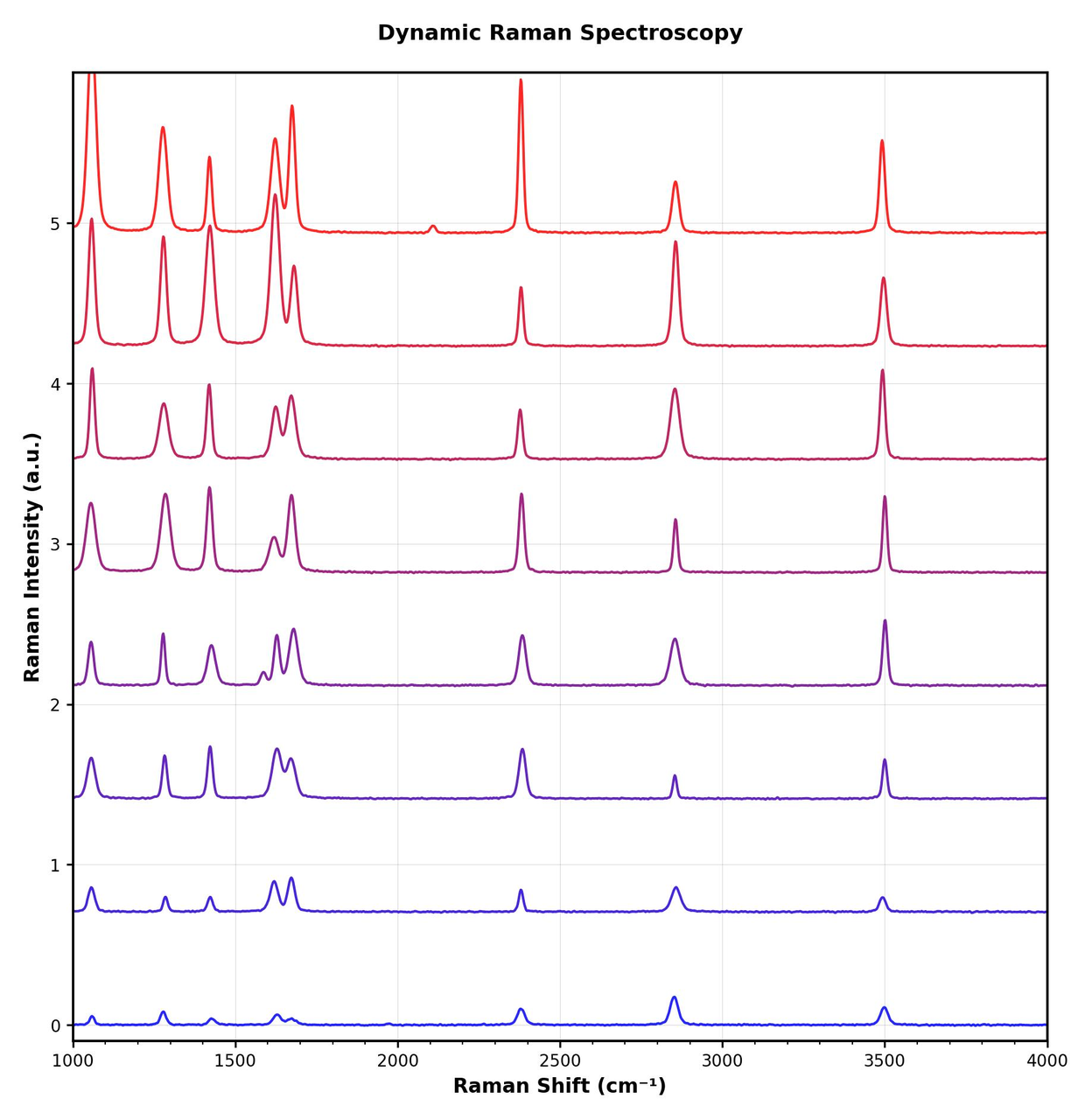}
    \subcaption{raman-generated}
\end{minipage}

\caption{Comparison between real spectral images and generated spectral images. The first row shows real images, while the second row shows generated spectral images of the same categories.}
\label{fig:overall}
\end{figure}

Based on the 20k selected real spectral images, we generated a total of 200k corresponding image datasets across the seven categories. We also designed QA pairs that samples key points from these images, which served as the training set for the underlying task. The prompt format for the QA pairs is shown in the Appendix, and a comparison between the generated data and the real data is shown in Figure \ref{fig:overall}. 
Specifically, we first generated the data based on the style and format of 20k real spectra. We then continuously optimized the generation method based on feedback from human scientists, iterating through versions until it was approved by them. Next, we generated batches of 10k data points each, from which human experts randomly selected $10\%$ for quality control. A batch was accepted only if its pass rate exceeded $95\%$; otherwise, it was discarded. This procedure was repeated until 20 qualified batches were obtained.

\section{Experiments and Results}
\label{sec:experiments}

\subsection{Experimental Setup}

We conduct a large-scale evaluation on the following set of several prominent MLLMs: Gemini-3-Flash-Preview \cite{gemini3flash}, Gemini-3-Pro-Preview \cite{gemini3pro}, Claude-4.5-Sonnet \cite{claude45sonnet}, Claude-4-Sonnet \cite{claude4sonnet}, Doubao-seed-1-6-flash \cite{doubao16flash}, DeepSeek-VL2 \cite{wu2024deepseek}, Qwen3-VL-8B Thinking \cite{bai2025qwen3vltechnicalreport}, Qwen3-VL-4B-Instruct \cite{bai2025qwen3vltechnicalreport}, Qwen3-VL-32B-Instruct \cite{bai2025qwen3vltechnicalreport}, GPT-4o \cite{openai2024gpt4o}, GPT-o4-mini \cite{openai2024gpt4mini}, GPT-5 \cite{openai_gpt5_2025}, GPT-o3 \cite{openai_o3_o4mini_2025}, Gemini-2.5-Pro \cite{deepmind_gemini_pro}, and Gemini-2.5-Flash \cite{google_gemini_flash}, and the Qwen3-VL-4B Backbone used for our method.
Our proposed model variants are \textbf{ours without the sampling strategy} (Qwen3-VL-4B trained on 20k QA data and 200k original underlying data, serving as the absolute baseline) and \textbf{ours with the sampling strategy} (Qwen3-VL-4B trained on 20k QA data and 200k sampling underlying data). The benchmark data does not overlap with the training set.

This experimental phase of our proposed model is supervised fine-tuning (SFT), designed to adapt the selected visual-language backbone model to the domain of our constructed SpecVQA benchmark, thereby improving the model's scientific understanding and quantification capabilities in spectral image question answering tasks. By selectively freezing the vision tower and multimodal projector, the fine-tuning strategy preserves the robustness and structural coherence of the pretrained visual representations, thereby promoting stable convergence and enhancing the model’s ability to capitalize on its established visual priors during downstream adaptation.

\subsection{Evaluation Metrics}
We used the following evaluation metrics in the two types of experimental tasks respectively.\\
\textbf{Scientific Accuracy.} We report results separately for L0 (Descriptive
Question) and L1 (Reasoning Question). Each answer is scored by GPT-o4-mini with reference to the standard answers. If the error rate falls within the predefined error tolerance of 5 percentage points, the answer is deemed correct; otherwise, it is regarded as incorrect. The accuracy rate is then calculated accordingly.\\
\textbf{Point Cloud Distance Score.} Used to quantify the fidelity of the data sampling and interpolation reconstruction strategy. We normalized it in actual application and compute the score as \eqref{score}. Here $d$ means the three kind of distance presented in Section 3.3.
\begin{equation}\label{score}
Score = 1 -\overline{d}\left(C_{\text {true }}, C_{\text {rec }}\right) 
= 1-\frac{{d}\left(C_{\text {true }}, C_{\text {rec }}\right)}{
\max _{x, y \in C_{\text {true }} \cup C_{\text {rec }}}\|x-y\|^2}.
\end{equation}
During the experiment, it was noticed that the number of lines output by some MLLMs differed from the original image. To address this, Hungarian algorithm \cite{kuhn1955hungarian} was employed to match the model's output with the real data, calculate the score for each line after matching, and then take the average of the valid data.

\subsection{SpecVQA Benchmark Leaderboard}

Performance evaluations of different MLLMs were conducted on our VQA benchmark. Table \ref{tab:leaderboard} presents the leaderboard, featuring the performance of MLLMs, with results clearly delineated across the two primary question categories. 

Ours model achieves the higher $\text{Sci-Acc}$ across both Information Extraction and Reasoning categories in both Chinese and English, demonstrating its superior ability to handle both fine-grained data retrieval and complex scientific deduction. This confirms that optimizing the input visual representation is significantly effective for this domain. Additionaly, it is clearly indicated that there is a substantial performance boost from our Data Sampling technique.

\begin{table*}[t]
\centering
\caption{Performance comparison of models on the SpecVQA benchmark.}
\label{tab:leaderboard}
\resizebox{\textwidth}{!}{
\begin{tabular}{|c|c|c|c|c|c|c|c|c|}
\hline
\multirow{2}{*}{Model} &
\multirow{2}{*}{Think} &
\multirow{2}{*}{Weight} &
\multirow{2}{*}{API-Version} &
\multicolumn{2}{c|}{English (en)} &
\multicolumn{2}{c|}{Chinese (zh)} &
\multirow{2}{*}{Overall} \\

\cline{5-8}

 &  &  & 
 & \makecell{Descriptive\\Question (L0)}
 & \makecell{Reasoning\\Question (L1)}
 & \makecell{Descriptive\\Question (L0)}
 & \makecell{Reasoning\\Question (L1)}
 &  \\

\hline

Gemini-3-Flash-Preview
& $\checkmark$
& Proprietary
& 20251217
& \textbf{0.7782}
& \textbf{0.7759}
& \textbf{0.8047}
& \textbf{0.7900}
& \textbf{0.7872} \\

Gemini-3-Pro-Preview
& $\checkmark$
& Proprietary
& 20251119
& 0.7718
& \textbf{0.7759}
& 0.7939
& 0.7731
& 0.7787 \\

Gemini-2.5-Pro
& $\checkmark$
& Proprietary
& 20250617
& 0.7645
& 0.7486
& 0.7758
& 0.7571
& 0.7615 \\

Gemini-2.5-Flash
& $\checkmark$
& Proprietary
& 20250617
& 0.7395
& 0.7043
& 0.7473
& 0.7185
& 0.7274 \\

GPT-5(high)
& $\checkmark$
& Proprietary
& 20250807
& 0.7017
& 0.6959
& 0.7115
& 0.7147 
& 0.7059 \\

GPT-o4mini
& $\checkmark$
& Proprietary
& 20250416
& 0.7144
& 0.6836
& 0.7237
& 0.6996
& 0.7054 \\

GPT-5(medium)
& $\checkmark$
& Proprietary
& 20250807
& 0.6923
& 0.6977
& 0.7154
& 0.7100
& 0.7039 \\

GPT-o3
& $\checkmark$
& Proprietary
& 20250416
& 0.6953
& 0.7100
& 0.7027
& 0.7043
& 0.7031 \\

GPT-5(low)
& $\checkmark$
& Proprietary
& 20250807
& 0.6825
& 0.6968
& 0.7066
& 0.6987
& 0.6961 \\

GPT-5.1
& ×
& Proprietary
& 20251113
& 0.6776
& 0.6347
& 0.6899
& 0.6290
& 0.6578 \\

GPT-5.2
& ×
& Proprietary
& 20251211
& 0.6776
& 0.6328
& 0.6899
& 0.6158
& 0.6540 \\

Claude-4.5-Sonnet
& $\checkmark$
& Proprietary
& 20250929
& 0.6148
& 0.5518
& 0.6315
& 0.5782
& 0.5941 \\

Doubao-seed-1-6-flash
& $\checkmark$
& Proprietary
& 20250828
& 0.6060
& 0.5687
& 0.6148
& 0.5574
& 0.5867 \\

Claude-4-Sonnet
& $\checkmark$
& Proprietary
& 20250514
& 0.5947
& 0.5282
& 0.5957
& 0.5565
& 0.5688 \\

Qwen3-VL-8B-Thinking
& $\checkmark$
& Open
& -
& 0.5864
& 0.5348
& 0.5805
& 0.5508
& 0.5631 \\

Doubao-seed-1-6-250615
& $\checkmark$
& Proprietary
& 20250615
& 0.5721
& 0.5499
& 0.5736
& 0.5452
& 0.5602 \\

Qwen3-VL-8B-Instruct
& ×
& Open
& -
& 0.5721
& 0.4256
& 0.5927
& 0.4727
& 0.5158 \\

Qwen3-Omni-30B-A3B-Thinking
& $\checkmark$
& Open
& -
& 0.5442
& 0.4868
& 0.5368
& 0.4765
& 0.5111 \\

Doubao-seed-1-6
& ×
& Proprietary
& 20250615
& 0.5530
& 0.4435
& 0.5697
& 0.4586
& 0.5062 \\

DeepSeek-VL2
& ×
& Open
& -
& 0.4657
& 0.3183
& 0.4092
& 0.3079
& 0.3753 \\

\hline
\end{tabular}
}
\end{table*}

\begin{table}[h]
\centering
\caption{Fidelity Comparison of the Data Sampling and Interpolation Reconstruction Strategy.}
\label{tab:chamfer_distance}
\resizebox{\textwidth}{!}{
\begin{tabular}{|c|c|c|c|c|}
\toprule
\textbf{Model} & \textbf{\makecell{Score-\\Chamfer Distance}} & \textbf{\makecell{Score-\\Hausdorff Distance}} & \textbf{\makecell{Score-\\Wasserstein Distance}}\\
\midrule
GPT-4o  & $0.3996$ & $0.3645$ & $0.4100$\\
Deepseek-VL2  & $0.4102$ & $0.3207$ & $0.4366$\\
Ours(without sampling strategy) & $0.4146$ &$0.3827$ &$0.4159$\\
Qwen3-VL-32B-Instruct & $0.7575$ & $0.7485$ & $0.7784$\\
Qwen3-VL-4B-Instruct &$0.7606$ &$0.7500$ &$0.7827$\\
GPT-o4-mini & $0.8229$ &$0.7819$ &$0.8413$\\
GPT-o3 & $0.8330$ &$0.7988$ &$0.8460$\\
GPT-5 & $0.8389$ &$0.7988$ &$0.8516$ \\
Gemini-2.5-Pro & $0.8893$ &$0.7042$ &$0.9051$\\
Gemini-2.5-Flash & $0.8953$ &$0.7994$ &$0.9109$\\
\textbf{Ours(with sampling strategy)} & $\mathbf{0.9776}$ &$\mathbf{0.9595}$ &$\mathbf{0.9810}$\\
\textbf{Testset with sampling strategy} &  $\mathbf{0.9899}$ &$\mathbf{0.9794}$ &$\mathbf{0.9904}$\\
\bottomrule
\end{tabular}
}
\end{table}

\subsection{Efficiency and Fidelity Analysis of Data Sampling strategy}

This section rigorously evaluates the impact of our data strategy in terms of input fidelity and final performance. The result is showed in Table \ref{tab:chamfer_distance}.

\subsubsection{Fidelity Evaluation.}
We first assess the quality of the point reconstruction using three distance scores. Specifically, we evaluate the fidelity of the reconstructed curve ($C_{\text{rec}}$) obtained by our approach relative to the original real data curve. The results indicate a high reconstruction accuracy, with a score approaching 0.99, which is shown as model line Testset with sampling strategy.

During experiments, we observe that most MLLMs perform a truncation operation on the input data. Some models such as Deepseek-VL2 handle excessively long tokens by directly replacing the remaining content with an ellipsis, rather than truncating or interrupting the processing. Our strategy substantially reduce visual token length to 6.7 percentage of the original, improving training efficiency and avoiding the cut-off of long data while preserving spectral fidelity.

\subsubsection{Ablation Study on Performance and Efficiency.}
Then we compare the performance of our model (ours with the sampling strategy) against the baseline (ours without the sampling strategy) and other MLLMs. 
The higher score for our strategy validates that our adaptive sampling strategy substantially preserves the spatial configuration and characteristic features of the original high-resolution spectrum, which is essential for accurate scientific analysis.

\section{Conclusion}
\label{sec:conclusion}

This work introduces SpecVQA, a comprehensive benchmark dedicated to evaluating the scientific visual intelligence of Multimodal Large Language Models across the seven most critical types of scientific spectra. By focusing on identifying and categorizing critical cases where general MLLMs fail, SpecVQA provides a necessary benchmark for advancing scientific MLLM research. Furthermore, our proposed Efficient Data Sampling and Interpolation Reconstruction strategy effectively addresses the inherent token length crisis associated with high-resolution spectral data, significantly improving the efficiency and feasibility of training domain-specific models. The establishment of a comprehensive leaderboard based on SpecVQA offers a clear and quantitative measure of current MLLM performance, guiding future research toward building truly expert-level scientific visual systems. In the future, we will further unify the two parts of the benchmark data and build professional scientific image datasets in a wider range of fields.

\par\vfill\par

\clearpage  


%
%
\bibliographystyle{splncs04}
\bibliography{main}

\newpage
\title{Appendix}
\author{}
\institute{}
\maketitle

\section{Cases in spectral image understanding by MLLMs}

In spectral image understanding, MLLMs typically target two complementary tasks: Scientific QA Task and Underlying Construction Task. Here we show some bad cases of these two tasks.

\subsection{Mislocalization of Fine Features
}

Inability to accurately locate or distinguish subtle spectral details, such as minor impurity peaks, small chemical shifts, or the precise positions of crystalline reflections.

\begin{figure}[h]
    \centering
    \includegraphics[width=0.5\textwidth]{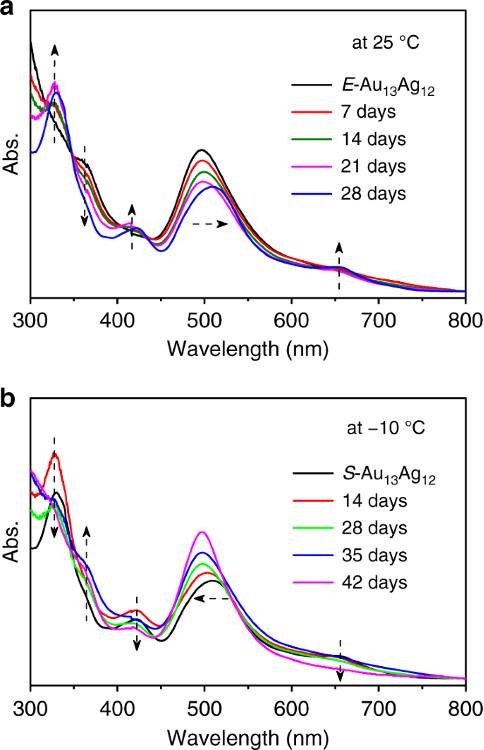}
        \caption{Case 1}
\end{figure}

\textbf{Q:} In panel a, how does the bandwidth of the absorption peak at approximately 350 nm change over time (from day 0 to day 28)?

\textbf{A:} Remains essentially unchanged.

\textbf{GPT-5:} Decreases .

\textbf{Gemini-2.5-pro:} Increases.

\textbf{Qwen3-VL-32B-Instruct:} Not visible.

\subsection{Inaccurate Quantitative Extraction
}

Errors in reading key numerical values from axes or curves, including integration ratios, maximum absorption wave lengths ($\lambda_{max}$), and precise binding energies.

\begin{figure}[h]
    \centering
    \includegraphics[width=0.5\textwidth]{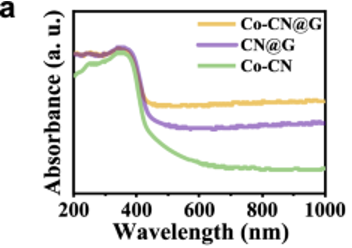}
        \caption{Case 2}
\end{figure}

\textbf{Q:} In Figure a, at approximately which wavelength is the absorption edge of the Co-CN sample (green curve) located?

\textbf{A:} Approximately 380 nm.

\textbf{GPT-5:} 450 nm.

\textbf{Gemini-2.5-pro:} 500 nm.

\textbf{Qwen3-VL-32B-Instruct:} Not visible.

\subsection{Flawed Mechanistic Reasoning
}

Logical inconsistencies when connecting visual evidence to scientific principles, resulting in incorrect structure elucidation, phase identification, or interpretation of fragmentation pathways.

\begin{figure}[h]
    \centering
    \includegraphics[width=0.5\textwidth]{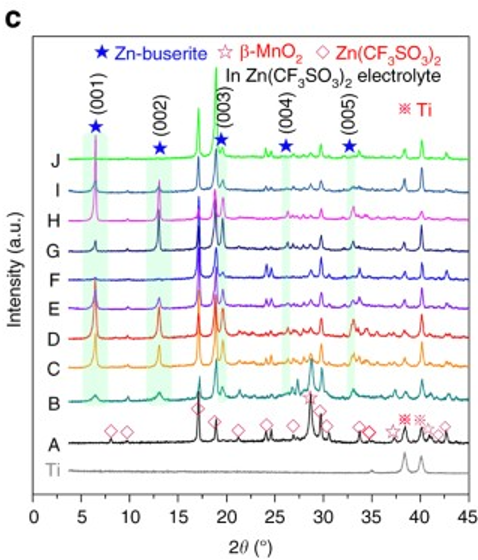}
        \caption{Case 3}
\end{figure}

\textbf{Q:} What are the primary diffraction peaks in the X-ray diffraction (XRD) pattern, and to which material do these peaks correspond??

\textbf{A:} The major diffraction peaks correspond to $\beta$-MnO$_2$ and Zn-buserite.

\textbf{GPT-5:} $\beta$-MnO$_2$.

\textbf{Gemini-2.5-pro:} Zn-buserite, $\beta$-MnO$_2$, Zn(CF$_3$SO$_3$)$_2$, InZn(CF$_3$SO$_3$)$_2$ electrolyte, Ti.

\textbf{Qwen3-VL-32B-Instruct:} (001), (002), (003), (004), (005) – Zn-buserite.

\subsection{Underlying Task}

Whether learning from or predicting these points, MLLMs will tokenize them into a longer sequence compared with those in other tasks, which might impose heavy computational and memory burdens. 

\begin{figure}[h]
    \centering
    \includegraphics[width=0.425\textwidth]{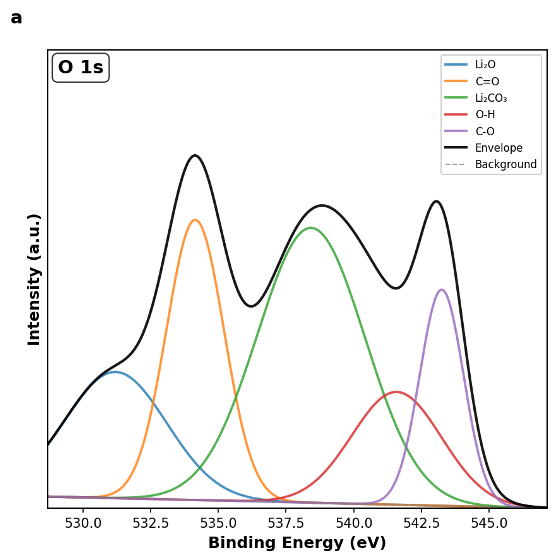}
        \caption{Case 4}
\end{figure}

\vspace{30pt}

\textbf{Deepseek-VL2:} 
\begin{Verbatim}[breaklines=true,breakanywhere=true]
<subplot A>
<line 1>[530.00,0.12],[532.50,0.30],...,[542.50,1.80],[545.00,0]</line>
<line 2>[530.00,0.00],[532.50,0.00],...,[542.50,0.00],[545.00,0.00]</line>
<line 3>[530.00,0.00],[532.50,0.00],...,[542.50,0.00],[545.00,0.00]</line>
<line 4>[530.00,0.00],[532.50,0.00],...,[542.50,0.00],[545.00,0.00]</line>
<line 5>[530.00,0.00],[532.50,0.00],...,[542.50,0.00],[545.00,0.00]</line>
<line 6>[530.00,0.00],[532.50,0.00],...,[542.50,0.00],[545.00,0.00]</line>
<line 7>[530.00,0.00],[532.50,0.00],...,[542.50,0.00],[545.00,0.00]</line>
<line 8>[530.00,0.00],[532.50,0.00],...,[542.50,0.00],[545.00,0.00]</line>
<line 9>[530.00,0.00],[532.50,0.00],...,[542.50,0.00],[545.00,0.00]</line>
</subplot>
\end{Verbatim}

\textbf{GPT-o4-mini:} 
\begin{Verbatim}[breaklines=true,breakanywhere=true]
<subplot A>
<line1>[528.00,0.00],[530.00,0.60],[532.00,0.10],[534.00,0.00],[536.00,0.00],[538.00,0.00],[540.00,0.00],[542.00,0.00],[544.00,0.00],[546.00,0.00]</line>
<line2>[528.00,0.00],[530.00,0.00],[532.00,0.80],[534.00,1.60],[536.00,0.50],[538.00,0.00],[540.00,0.00],[542.00,0.00],[544.00,0.00],[546.00,0.00]</line>
<line3>[528.00,0.00],[530.00,0.00],[532.00,0.00],[534.00,0.30],[536.00,1.80],[538.00,2.00],[540.00,1.80],[542.00,0.80],[544.00,0.10],[546.00,0.00]</line>
<line4>[528.00,0.00],[530.00,0.00],[532.00,0.00],[534.00,0.00],[536.00,0.00],[538.00,0.10],[540.00,0.30],[542.00,0.80],[544.00,0.50],[546.00,0.00]</line>
<line5>[528.00,0.00],[530.00,0.00],[532.00,0.00],[534.00,0.00],[536.00,0.00],[538.00,0.00],[540.00,0.00],[542.00,0.10],[544.00,0.60],[546.00,0.10]</line>
<line6>[528.00,0.05],[530.00,0.65],[532.00,0.95],[534.00,1.95],[536.00,2.35],[538.00,2.15],[540.00,2.15],[542.00,1.75],[544.00,1.25],[546.00,0.15]</line>
<line7>[528.00,0.05],[530.00,0.05],[532.00,0.05],[534.00,0.05],[536.00,0.05],[538.00,0.05],[540.00,0.05],[542.00,0.05],[544.00,0.05],[546.00,0.05]</line>
</subplot>
\end{Verbatim}

\textbf{GPT-5:} 
\begin{Verbatim}[breaklines=true,breakanywhere=true]
<subplot A>\n<line 1>[529.00,0.30],[531.00,0.80],[533.00,0.30],[535.00,0.05],[537.00,0.01],[539.00,0.00],[541.00,0.00],[543.00,0.00],[545.00,0.00]</line>\n<line 2>[529.00,0.02],[531.00,0.10],[533.00,1.10],[535.00,1.30],[537.00,0.20],[539.00,0.03],[541.00,0.00],[543.00,0.00],[545.00,0.00]</line>\n<line 3>[529.00,0.00],[531.00,0.00],[533.00,0.05],[535.00,0.20],[537.00,1.00],[539.00,1.10],[541.00,0.40],[543.00,0.05],[545.00,0.00]</line>\n<line 4>[529.00,0.00],[531.00,0.00],[533.00,0.00],[535.00,0.05],[537.00,0.10],[539.00,0.20],[541.00,0.50],[543.00,0.40],[545.00,0.10]</line>\n<line 5>[529.00,0.00],[531.00,0.00],[533.00,0.00],[535.00,0.00],[537.00,0.05],[539.00,0.10],[541.00,0.60],[543.00,1.10],[545.00,0.30]</line>\n<line 6>[529.00,0.42],[531.00,0.99],[533.00,1.53],[535.00,1.67],[537.00,1.42],[539.00,1.48],[541.00,1.54],[543.00,1.58],[545.00,0.42]</line>\n<line 7>[529.00,0.10],[531.00,0.09],[533.00,0.08],[535.00,0.07],[537.00,0.06],[539.00,0.05],[541.00,0.04],[543.00,0.03],[545.00,0.02]</line>\n</subplot>
\end{Verbatim}

\textbf{Gemini-2.5-pro:} 
\begin{Verbatim}[breaklines=true,breakanywhere=true]
<subplot A>\n<line 1>[530.13,0.09],[530.51,0.21],[530.93,0.44],[531.41,0.69],[531.75,0.81],[532.22,0.80],[532.68,0.58],[533.15,0.33],[533.64,0.15],[534.12,0.06],[534.60,0.03]</line>\n<line 2>[531.75,0.05],[532.32,0.14],[532.84,0.35],[533.32,0.70],[533.79,1.17],[534.21,1.48],[534.69,1.43],[535.15,1.05],[535.64,0.60],[536.14,0.28],[536.60,0.10],[537.09,0.03]</line>\n<line 3>[533.36,0.02],[534.40,0.06],[535.27,0.18],[536.10,0.50],[536.87,1.05],[537.38,1.42],[537.84,1.58],[538.35,1.49],[538.93,1.13],[539.52,0.71],[540.10,0.40],[540.67,0.20],[541.24,0.09],[541.83,0.04],[542.42,0.01]</line>\n<line 4>[538.45,0.02],[539.11,0.05],[539.75,0.15],[540.36,0.33],[540.88,0.54],[541.34,0.65],[541.80,0.62],[542.33,0.46],[542.86,0.29],[543.39,0.15],[543.92,0.06],[544.47,0.02]</line>\n<line 5>[540.23,0.01],[540.82,0.06],[541.36,0.21],[541.88,0.49],[542.35,0.83],[542.72,1.04],[543.18,0.99],[543.66,0.69],[544.15,0.38],[544.62,0.16],[545.00,0.06]</line>\n

<line 6>[529.98,0.36],[530.51,0.55],[531.02,0.77],[531.55,0.96],[532.06,1.12],[532.57,1.29],[533.06,1.48],[533.53,1.68],[534.00,1.79],[534.46,1.82],[534.98,1.69],[535.53,1.39],[536.03,1.13],[536.52,1.11],[537.02,1.27],[537.54,1.49],[538.03,1.59],[538.53,1.52],[539.04,1.29],[539.54,1.01],[540.03,0.79],[540.54,0.77],[541.04,0.91],[541.55,1.17],[542.06,1.41],[542.54,1.58],[542.92,1.64],[543.41,1.42],[543.90,0.99],[544.40,0.57],[544.89,0.29]</line>\n<line 7>[530.00,0.08],[532.50,0.06],[535.00,0.05],[537.50,0.04],[540.00,0.03],[542.50,0.01],[545.00,0.00]</line>\n</subplot>
\end{Verbatim}

\textbf{Qwen3-VL-32B-Instruct:} 
\begin{Verbatim}[breaklines=true,breakanywhere=true]
<subplot a>\n<line 1>[530.00,0.00],[530.50,0.15],[531.00,0.30],[531.50,0.45],[532.00,0.60],[532.50,0.75],[533.00,0.85],[533.50,0.90],[534.00,0.95],[534.50,1.00],[535.00,1.05],[535.50,1.08],[536.00,1.10],[536.50,1.12],[537.00,1.13],[537.50,1.14],[538.00,1.15],[538.50,1.15],[539.00,1.15],[539.50,1.15],[540.00,1.15],[540.50,1.15],[541.00,1.15],[541.50,1.15],[542.00,1.15],[542.50,1.15],[543.00,1.15],[543.50,1.15],[544.00,1.15],[544.50,1.15],[545.00,1.15],[545.50,1.15],[546.00,1.15],[546.50,1.15],[547.00,1.15],[547.50,1.15],[548.00,1.15],[548.50,1.15],[549.00,1.15],[549.50,1.15],[550.00,1.15],[550.50,1.15],[551.00,1.15],[551.50,1.15],[552.00,1.15],[552.50,1.15],[553.00,1.15],[553.50,1.15],[554.00,1.15],[554.50,1.15],[555.00,1.15],[555.50,1.15],[556.00,1.15],[556.50,1.15],[557.00,1.15],[557.50,1.15],[558.00,1.15],[558.50,1.15],[559.00,1.15],[559.50,1.15],[560.00,1.15],[560.50,1.15],[561.00,1.15],[561.50,1.15],[562.00,1.15],[562.50,1.15],[563.00,1.15],[563.50,1.15],[564.00,1.15],[564.50,1.15],[565.00,1.15],[565.50,1.15],[566.00,1.15],[566.50,1.15],[567.00,1.15],[567.50,1.15],[568.00,1.15],[568.50,1.15],[569.00,1.15],[569.50,1.15],[570.00,1.15],[570.50,1.15],[571.00,1.15],[571.50,1.15],[572.00,1.15],[572.50,1.15],[573.00,1.15],[573.50,1.15],[574.00,1.15],[574.50,1.15],[575.00,1.15],[575.50,1.15],[576.00,1.15],[576.50,1.15],[577.00,1.15],[577.50,1.15],[578.00,1.15],[578.50,1.15],[579.00,1.15],[579.50,1.15],[580.00,1.15],[580.50,1.15],[581.00,1.15],[581.50,1.15],[582.00,1.15],[582.50,1.15],[583.00,1.15],[583.50,1.15],[584.00,1.15],[584.50,1.15],[585.00,1.15],[585.50,1.15],[586.00,1.15],[586.50,1.15],[587.00,1.15],[587.50,1.15],[588.00,1.15],[588.50,1.15],[589.00,1.15],[589.50,1.15],[590.00,1.15],[590.50,1.15],[591.00,1.15],[591.50,1.15],[592.00,1.15],[592.50,1.15],[593.00,1.15],[593.50,1.15],[594.00,1.15],[594.50,1.15],[595.00,1.15],[595.50,1.15],[596.00,1.15],[596.50,1.15],[597.00,1.15],[597.50,1.15],[598.00,1.15],[598.50,1.15],[599.00,1.15],[599.50,1.15],[600.00,1.15]</line>\n

<line 2>[530.00,0.00],[530.50,0.02],[531.00,0.05],[531.50,0.09],[532.00,0.14],[532.50,0.21],[533.00,0.30],[533.50,0.40],[534.00,0.51],[534.50,0.62],[535.00,0.73],[535.50,0.83],[536.00,0.92],[536.50,1.00],[537.00,1.06],[537.50,1.10],[538.00,1.12],[538.50,1.13],[539.00,1.13],[539.50,1.12],[540.00,1.10],[540.50,1.07],[541.00,1.03],[541.50,0.98],[542.00,0.92],[542.50,0.85],[543.00,0.78],[543.50,0.71],[544.00,0.64],[544.50,0.57],[545.00,0.51],[545.50,0.45],[546.00,0.40],[546.50,0.35],[547.00,0.32],[547.50,0.29],[548.00,0.27],[548.50,0.25],[549.00,0.23],[549.50,0.22],[550.00,0.20],[550.50,0.19],[551.00,0.18],[551.50,0.17],[552.00,0.16],[552.50,0.16],[553.00,0.15],[553.50,0.15],[554.00,0.15],[554.50,0.15],[555.00,0.15],[555.50,0.15],[556.00,0.15],[556.50,0.15],[557.00,0.15],[557.50,0.15],[558.00,0.15],[558.50,0.15],[559.00,0.15],[559.50,0.15],[560.00,0.15],[560.50,0.15],[561.00,0.15],[561.50,0.15],[562.00,0.15],[562.50,0.15],[563.00,0.15],[563.50,0.15],[564.00,0.15],[564.50,0.15],[565.00,0.15],[565.50,0.15],[566.00,0.15],[566.50,0.15],[567.00,0.15],[567.50,0.15],[568.00,0.15],[568.50,0.15],[569.00,0.15],[569.50,0.15],[570.00,0.15],[570.50,0.15],[571.00,0.15],[571.50,0.15],[572.00,0.15],[572.50,0.15],[573.00,0.15],[573.50,0.15],[574.00,0.15],[574.50,0.15],[575.00,0.15],[575.50,0.15],[576.00,0.15],[576.50,0.15],[577.00,0.15],[577.50,0.15],[578.00,0.15],[578.50,0.15],[579.00,0.15],[579.50,0.15],[580.00,0.15],[580.50,0.15],[581.00,0.15],[581.50,0.15],[582.00,0.15],[582.50,0.15],[583.00,0.15],[583.50,0.15],[584.00,0.15],[584.50,0.15],[585.00,0.15],[585.50,0.15],[586.00,0.15],[586.50,0.15],[587.00,0.15],[587.50,0.15],[588.00,0.15],[588.50,0.15],[589.00,0.15],[589.50,0.15],[590.00,0.15],[590.50,0.15],[591.00,0.15],[591.50,0.15],[592.00,0.15],[592.50,0.15],[593.00,0.15],[593.50,0.15],[594.00,0.15],[594.50,0.15],[595.00,0.15],[595.50,0.15],[596.00,0.15],[596.50,0.15],[597.00,0.15],[597.50,0.15],[598.00,0.15],[598.50,0.15],[599.00,0.15],[599.50,0.15],[600.00,0.15]</line>\n<line 3>[530.00,0.00],[530.50,0.01],[531.00,0.03],[531.50,0.06],[532.00,0.09],[532.50,0.14],[533.00,0.19],[533.50,0.25],[534.00,0.31],[534.50,0.37],[535.00,0.43],[535.50,0.48],[536.00,0.53],[536.50,0.57],[537.00,0.60],[537.50,0.63],[538.00,0.65],[538.50,0.66],[539.00,0.67],[539.50,0.67],[540.00,0.66],[540.50,0.65],[541.00,0.64],[541.50,0.62],[542.00,0.60],[542.50,0.57],[543.00,0.54],[543.50,0.50],[544.00,0.46],[544.50,0.41],[545.00,0.37],[545.50,0.32],[546.00,0.28],[546.50,0.24],[547.00,0.21],[547.50,0.18],[548.00,0.16],[548.50,0.14],[549.00,0.12],[549.50,0.11],[550.00,0.10],[550.50,0.09],[551.00,0.08],[551.50,0.08],[552.00,0.07],[552.50,0.07],[553.00,0.07],[553.50,0.07],[554.00,0.07],[554.50,0.07],[555.00,0.07],[555.50,0.07],[556.00,0.07],[556.50,0.07],[557.00,0.07],[557.50,0.07],[558.00,0.07],[558.5"\
\end{Verbatim}

\section{Qualitative analysis of benchmarks}

Table~\ref{tab:benchmark_comparison} presents a qualitative comparison between SpecVQA and existing visual question answering benchmarks. Compared with prior datasets that mainly focus on general charts or synthetic plots, SpecVQA specifically targets spectral figures from scientific literature and introduces underlying scientific tasks.

\begin{table}[h]
\centering
\caption{Comparison of SpecVQA with existing visual question answering benchmarks.}
\label{tab:benchmark_comparison}
\footnotesize

\begin{tabular}{
>{\centering\arraybackslash}m{1.8cm}
>{\centering\arraybackslash}m{2.2cm}
>{\centering\arraybackslash}m{2.1cm}
>{\centering\arraybackslash}m{2.6cm}
>{\centering\arraybackslash}m{1.4cm}
>{\centering\arraybackslash}m{1.6cm}
}

\toprule
Benchmark & Domain & Scale & QA Types & Answer & Underlying \\
\midrule

FigureQA & Synthetic plots & ~180K/~2.3M & Comparison, color matching & Binary &  $\times$ \\

DVQA & Bar charts & 300K / 3.4M & Counting, value lookup & Num/Text &  $\times$ \\

PlotQA & Scientific plots & 224K/28.9M & Data extraction, comparison & Numeric & $\times$ \\

ChartQA & Chart figures & 21.9K/32.7K & Reasoning, aggregation & Num/Text & $\times$ \\

SciVQA & Scientific diagrams & $\sim$10K/$\sim$50K & Scientific reasoning & Text & $\times$ \\

EXAMS-V & Exam figures & $\sim$24K/$\sim$100K & Knowledge QA & Multi-choice & $\times$ \\

\textbf{SpecVQA} & Spectral figures &  620/3100 & L0: Descriptive; L1: Reasoning & Num/Text & $\checkmark$ \\

\bottomrule
\end{tabular}

\end{table}

\section{Labeling Guidelines}
To ensure the scientific validity and representativeness of the benchmark, a PhD team of domain experts manually curated a subset of 620 spectral figures from 20k candidates collected from peer-reviewed journals and open-access scientific databases. The selection process was guided by six key criteria: Spectra Type, Image Structure, Text Completeness, Subplot Correlation, Sample Diversity and Resolution.

For each figure, domain experts carefully designed 5 Question-Answer(QA) pairs. We expect the questions and answers to reflect issues that are genuinely of interest in their research and pose challenges to MLLMs, rather than trivial or nonsensical visual QAs. Therefore, these pairs were further refined through multiple rounds of rigorous review and revision to ensure both clarity and scientific accuracy.

The final 3,100 QA-pairs are classified into two critical categoriesbased on the required cognitive effort. We provided both Chinese version and English version to test the scientific performance of models in different languages.

\textbf{Category 1 (L0): Descriptive Question} 

This category focuses on the ability to directly understand visual information, including:
\begin{itemize}
    \item Information extraction: extracting titles, labels, legends and x/y-axis information.
    \item Value localization: locating maximum or minimum points (e.g., strongest peaks) and peak location or range.
    \item Pattern recognition: identifying entities that meet specific conditions.
    \item Layout understanding: analyzing multi-panel subplots.
    \item Classification: classifying features (e.g., peak shapes).
\end{itemize}

\textbf{Category 2 (L1): Reasoning Question}

This category emphasizes the ability to analyze and reason based on image content, including:
\begin{itemize}
    \item Comparison: comparing multiple entities and drawing conclusions.
    \item Counting: determining the number of elements that satisfy certain conditions.
    \item Calculation: performing computations on data presented in the figure.
    \item Trend analysis: predicting changes in peak shapes or trends.
    \item Causal analysis: analyzing and understanding the scientific problems reflected in the figure.
\end{itemize}

The following is a complete annotation guide for industry experts:

For each image, we pre-generated 5 QA pairs using large models such as GPT. All QA pairs are open-ended, and the answers are as concise as possible; that is, the answer may be a number or a phrase, and sentences should be avoided unless absolutely necessary.
For each image, the order of work for each expert is as follows:

\begin{itemize}
    \item Determine the image itself: Analyze whether the current image has analytical value. If the result is "meaningful," proceed to the next step; if the result is "meaningless," clear all 5 QA pairs for the current image (keeping the original image), indicating that the current image is meaningless. This step is to prevent the inclusion of some worthless or low-value images that contaminate the data. A meaningful spectrum should include: horizontal and vertical axes and their corresponding units, necessary legends or symbols, relatively complete peaks/signals, and a clear image.
    \item Evaluate each of the 5 QA pairs individually: First, determine if the current Q's question aligns with the current major and the current image. If any mismatch is found, modify the Q itself. If partial modifications still do not meet the requirements, delete the current Q and rewrite it. Finally, determine if the current A's answer satisfies the issues just corrected. If not, modify or rewrite the current A until it does.
\end{itemize}

\textbf{Tips}
\begin{itemize}
    \item Due to the limitations of Uniparser's parsing capabilities, some images may have incorrect or incomplete slices. Encountering such issues is essentially meaningless.
    \item To reflect the capabilities of large-scale models in real-world applications as objectively as possible, we encourage experts to design questions that are more closely aligned with actual research needs. Any questions and answers that can be derived from the graphs are welcome and can be documented; there's no need to strictly adhere to MLLMs settings, as MLLMs current research capabilities regarding spectra are limited. We encourage industry experts to take initiative and design appropriate assessment criteria. Questions that are particularly challenging for current large-scale models are especially welcome.
    \item If you encounter any other issues with the annotations, please feel free to contact us. Thank you for your valuable questions and suggestions.
\end{itemize}

\section{Prompt Examples}

\subsection{Generate QA pairs}

We use the following prompt template for question answering.\\

\textbf{Prompt:}

\begin{Verbatim}[
breaklines=true, 
breakanywhere=true, 
breaksymbol={},
breaksymbolleft={},
breaksymbolright={},
breaksymbolsepleft=0pt,
breaksymbolsepright=0pt
]
"""You will perform a Vision Question Answering (VQA) task based on a scientific literature image.
I will provide you with an image and a question related to it.

Requirements:

- Answer **only based on information directly visible in the image**, without relying on any prior knowledge or assumptions.

- Do **not infer, reason, or speculate**. Only describe content that can be directly observed.

- The answer must be **very concise**, preferably a **number, word, or short phrase**, and **not a full sentence**.

- If the information is not visible in the image, answer: **Not visible**.

Question:
{QUESTION}

Output **only the answer**, and do not include any other content:
"""
\end{Verbatim}

\subsection{Evaluate model performance on (VQA) tasks}

This benchmark evaluates model performance on Visual Question Answering (VQA) tasks. Following the ChartVLM, GPT-o4-mini serves solely as a judge to score the model’s predictions against the ground truth. A predefined error tolerance of 5 percentage points is applied: if the error falls within this range, the answer is considered correct; otherwise, it is marked incorrect. Accuracy is then computed based on these judgments. \footnote{https://github.com/Alpha-Innovator/ChartVLM/blob/main/eval/metric/gpt\_acc.py} \\

\textbf{Prompt:}

\begin{Verbatim}[
breaklines=true, 
breakanywhere=true, 
breaksymbol={},
breaksymbolleft={},
breaksymbolright={},
breaksymbolsepleft=0pt,
breaksymbolsepright=0pt
]
"""Given multiple question-answer pairs and the corresponding predictions, evaluate the correctness of predictions. The output should be only 'True' or 'False'. Note that if the groundtruth answer is a numeric value with/without the unit, impose 5 percentage error tolerance to the answer, e.g., the answer of 95 is marked as correct when groundtruth value is 100 million.

User: <question> What was the incremental increase in revenue from 2020 to 2021? <groundtruth answer> 5 million $ <answer> 20 </s>
A: False

User: <question> What percentage of government spending was allocated to infrastructure in 2020? <groundtruth answer> 10 percentage <answer> 14-4=10 </s>
A: True

User: <question> What is the total production of Wind Energy in the four months from January to April 2021? <groundtruth answer> 2300 MW <answer> The total production of Wind Energy in the four months from January to April 2021 is 2450 MW.
A: True

User: <question> What is the total of manufactured goods for UK and Germany combined? <groundtruth answer> 5 <answer> Five
A: True

User: <question> {QUESTION} <groundtruth answer> {GROUND TRUTH} <answer> {PREDICTION} </s>
AI: 
"""
\end{Verbatim}

\subsection{Extract data points in the underlying task}

For large, untuned models,  our prompts are as follows.\\

\textbf{Prompt:}

\begin{Verbatim}[
breaklines=true, 
breakanywhere=true, 
breaksymbol={},
breaksymbolleft={},
breaksymbolright={},
breaksymbolsepleft=0pt,
breaksymbolsepright=0pt
]
"""You will be given a composite scientific figure and a question.
Your task is to identify the specific subplot referenced in the question and extract the underlying data series from that subplot.

For every line in the selected subplot, output its extracted data points in the following strict format:

<subplot {{SUBPLOT_NAME}}>
<line 1>[x1,y1],[x2,y2],...[xn,yn]</line>
<line 2>[x1,y1],[x2,y2],...[xn,yn]</line>
...
</subplot>

Requirements:

- Output must follow the exact format above with no extra text.

- {{SUBPLOT_NAME}} will be a letter (A, B, C, ...) provided in the question.

- Each data series (line) must be enclosed in <line i></line>.

- Each (x,y) pair must be enclosed in brackets [x,y] and values should keep two decimal places.

- Do not add explanations, summaries, or comments outside the format.

Example Output (for illustration)
<subplot A>
<line 1>[0.00,1.25],[1.00,2.50],[2.00,3.75]</line>
<line 2>[0.00,0.80],[1.00,1.60],[2.00,2.40]</line>
</subplot>

Here is your question:
{QUESTION}
"""
\end{Verbatim}

For our two fine-tuned models, namely the Qwen3-VL-4B model with and without a sampling strategy, we can directly input the Question. Here are examples of the Question and Answer for our Underlying task.\\

\textbf{Prompt:}

\begin{Verbatim}[
breaklines=true, 
breakanywhere=true, 
breaksymbol={},
breaksymbolleft={},
breaksymbolright={},
breaksymbolsepleft=0pt,
breaksymbolsepright=0pt
]
{'conversations': [{'from': 'human',
   'value': '<image>Underlying data for subplot B:'},
  {'from': 'gpt',
   'value': '<subplot B><line 0>[393.00,0.12],[393.36,0.14]...[414.63,0.00],[415.59,0.00]</line><line 2>[393.00,0.12],[393.36,0.14]...[414.63,0.00],[415.59,0.00]</line></subplot>'}],
 'images': ['data20/199999.png']}
\end{Verbatim}

\section{SFT parameters}

\begin{Verbatim}[
breaklines=true, 
breakanywhere=true, 
breaksymbol={},
breaksymbolleft={},
breaksymbolright={},
breaksymbolsepleft=0pt,
breaksymbolsepright=0pt
]
### method
stage: sft
do_train: true
finetuning_type: full
freeze_vision_tower: true
freeze_multi_modal_projector: true
freeze_language_model: false
deepspeed: examples/deepspeed/ds_z3_config.json

template: qwen3_vl_nothink
cutoff_len: 16384

### train
per_device_train_batch_size: 2
gradient_accumulation_steps: 4
learning_rate: 1.0e-5
num_train_epochs: 1.0
lr_scheduler_type: cosine
warmup_ratio: 0.1
bf16: true
ddp_timeout: 180000000
resume_from_checkpoint: null
flash_attn: fa2
enable_liger_kernel: True
\end{Verbatim}

\end{document}